\documentclass[runningheads]{llncs}
\usepackage[T1]{fontenc}
\usepackage{multicol}
\usepackage{subcaption}
\usepackage{xcolor}
\usepackage{csquotes}
\usepackage{bbm}
\usepackage{bm}
\newcommand{\norm}[1]{\left\lVert#1\right\rVert}
\usepackage{amsmath}
\usepackage{hyperref}
\usepackage{bbm}
\usepackage[misc]{ifsym}

\usepackage{algorithm,algpseudocode}
\algnewcommand{\algorithmicforeach}{\textbf{for each}}
\algdef{SE}[FOR]{ForEach}{EndForEach}[1]
  {\algorithmicforeach\ #1\ \algorithmicdo}
  {\algorithmicend\ \algorithmicforeach}
\usepackage{graphicx}

\usepackage{color}

\usepackage{fancyhdr}

\fancypagestyle{mystyle}{%
    \lhead{Published as a conference paper at ECML PKDD 2022}\fancyfoot{}
} 
\begin{document}
\title{Avoiding Forgetting and Allowing Forward Transfer in Continual Learning via Sparse Networks}
\titlerunning{Avoiding Forgetting and Allowing Forward Transfer}
%
%
\institute{}
\author{Ghada Sokar\inst{1}{\Letter} \and
Decebal Constantin Mocanu\inst{2,1} \and
Mykola Pechenizkiy\inst{1}}
\authorrunning{G. Sokar et al.}
%
\institute{Eindhoven University of
Technology, Eindhoven, The Netherlands\\ 
\email{\{g.a.z.n.sokar, m.pechenizkiy\}@tue.nl} \and
University of Twente, Enschede, The Netherlands\\
\email{d.c.mocanu@utwente.nl}}

\maketitle              
\thispagestyle{mystyle}

\begin{abstract}
Using task-specific components within a neural network in continual learning (CL) is a compelling strategy to address the \textit{stability-plasticity} dilemma in fixed-capacity models without access to past data. Current methods focus only on selecting a sub-network for a new task that reduces forgetting of past tasks. However, this selection could limit the forward transfer of \textit{relevant} past knowledge that helps in future learning. Our study reveals that satisfying both objectives jointly is more challenging when a unified classifier is used for all classes of seen tasks--class-Incremental Learning (class-IL)--as it is prone to ambiguities between classes across tasks. Moreover, the challenge increases when the semantic similarity of classes across tasks increases. To address this challenge, we propose a new CL method, named AFAF\footnote{Code is available at: https://github.com/GhadaSokar/AFAF.}, that aims to Avoid Forgetting and Allow Forward transfer in class-IL using fix-capacity models. AFAF allocates a sub-network that enables \textit{selective} transfer of relevant knowledge to a new task while preserving past knowledge, \textit{reusing} some of the previously allocated components to utilize the fixed-capacity, and addressing class-ambiguities when similarities exist. The experiments show the effectiveness of AFAF in providing models with multiple CL desirable properties, while outperforming state-of-the-art methods on various challenging benchmarks with different semantic similarities.  

\end{abstract}
\thispagestyle{mystyle}
\section{Introduction}
\label{sec:intro}
\begin{figure*} [h]
  \centering
  \begin{subfigure}[b]{0.21\columnwidth}
  \centering
    \includegraphics[width=1.2\linewidth]{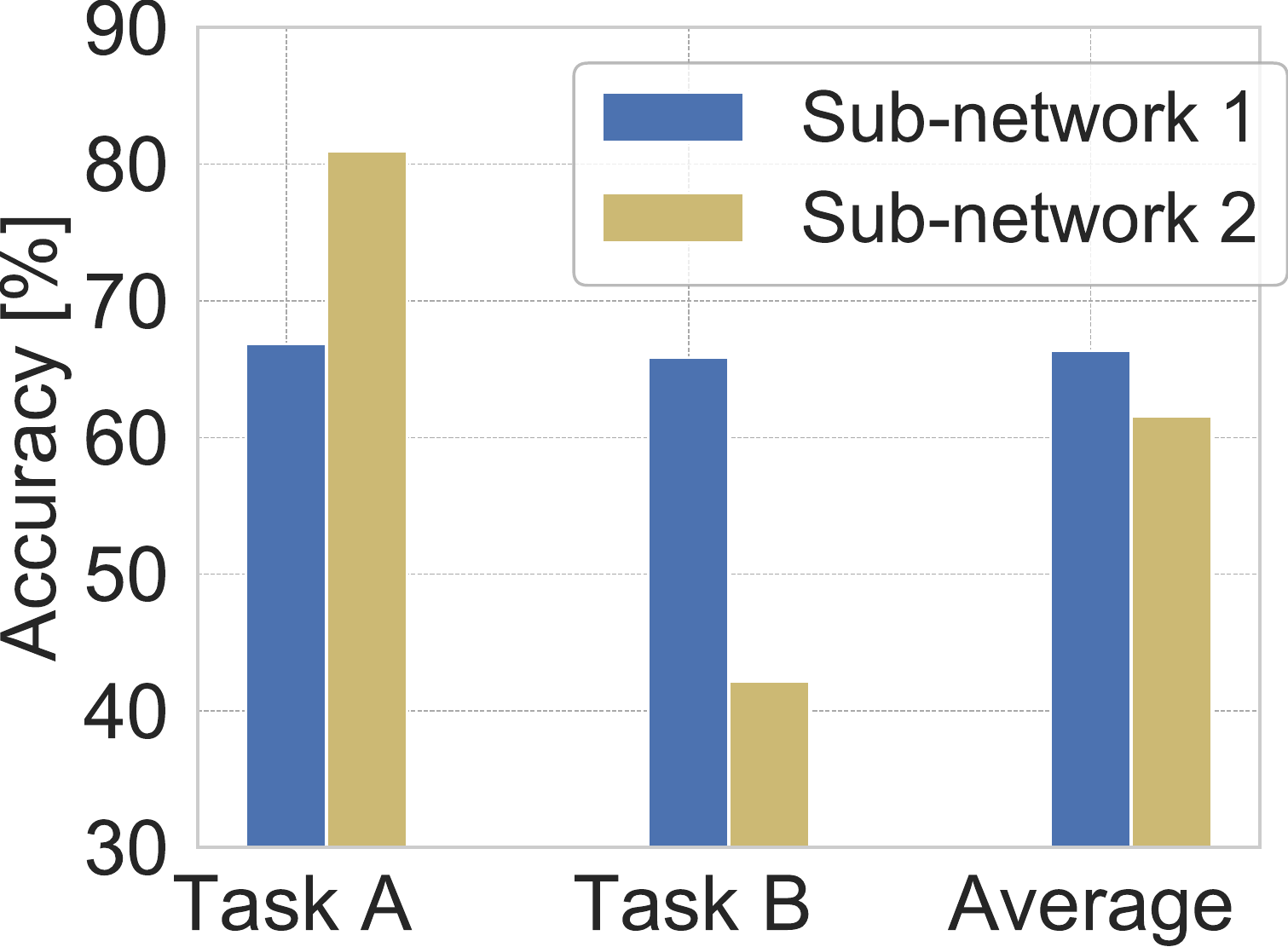}
    \caption{} 
    \label{fig:initial_topology}
  \end{subfigure}
    \hfill
    \begin{subfigure}[b]{0.21\columnwidth}
    \includegraphics[width=1.2\linewidth]{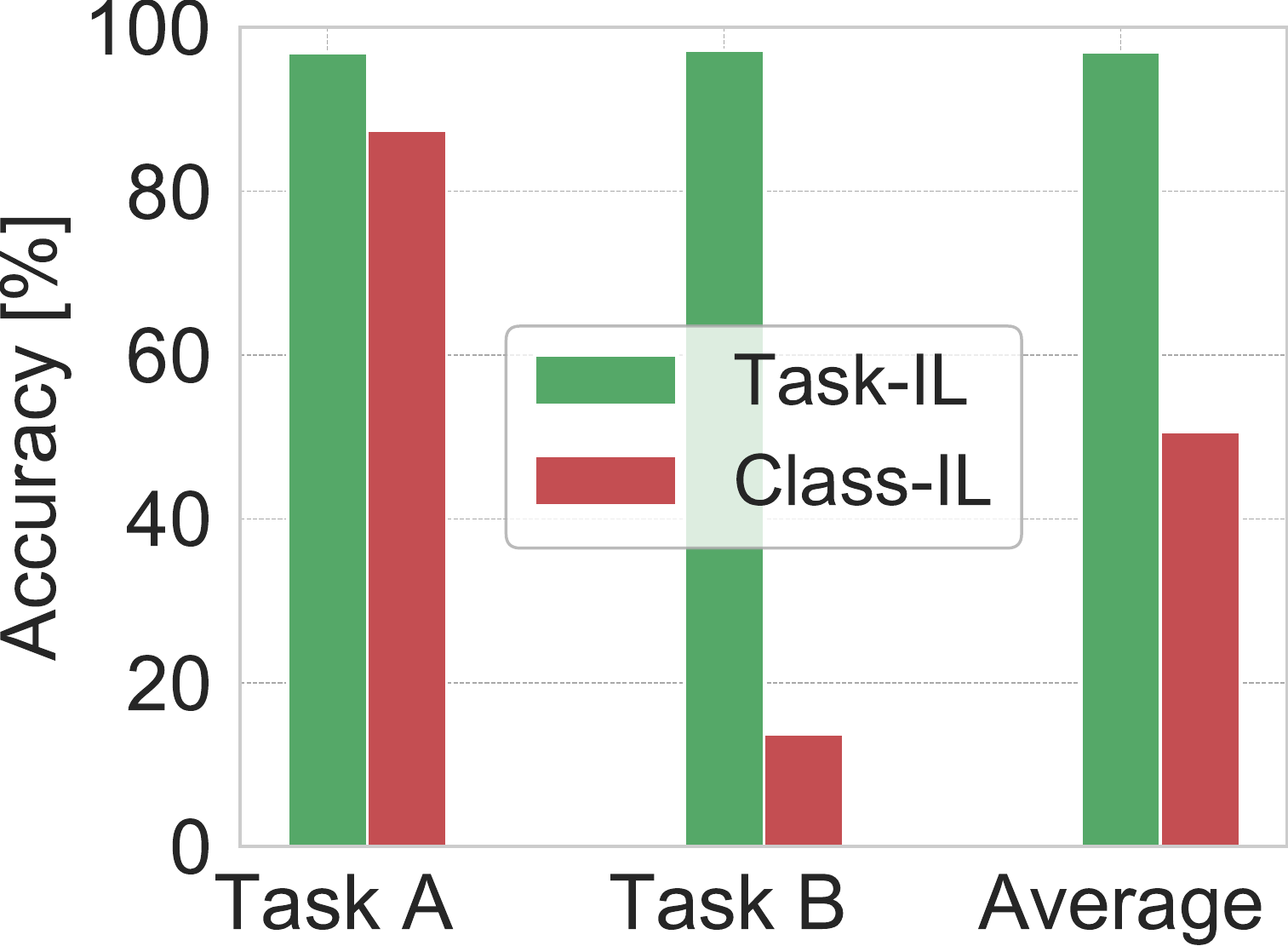}
    \caption{} 
    \label{fig:acc_TIL_CIL}
  \end{subfigure}
  \hfill
    \begin{subfigure}[b]{0.4\columnwidth}
    \includegraphics[width=0.5\linewidth]{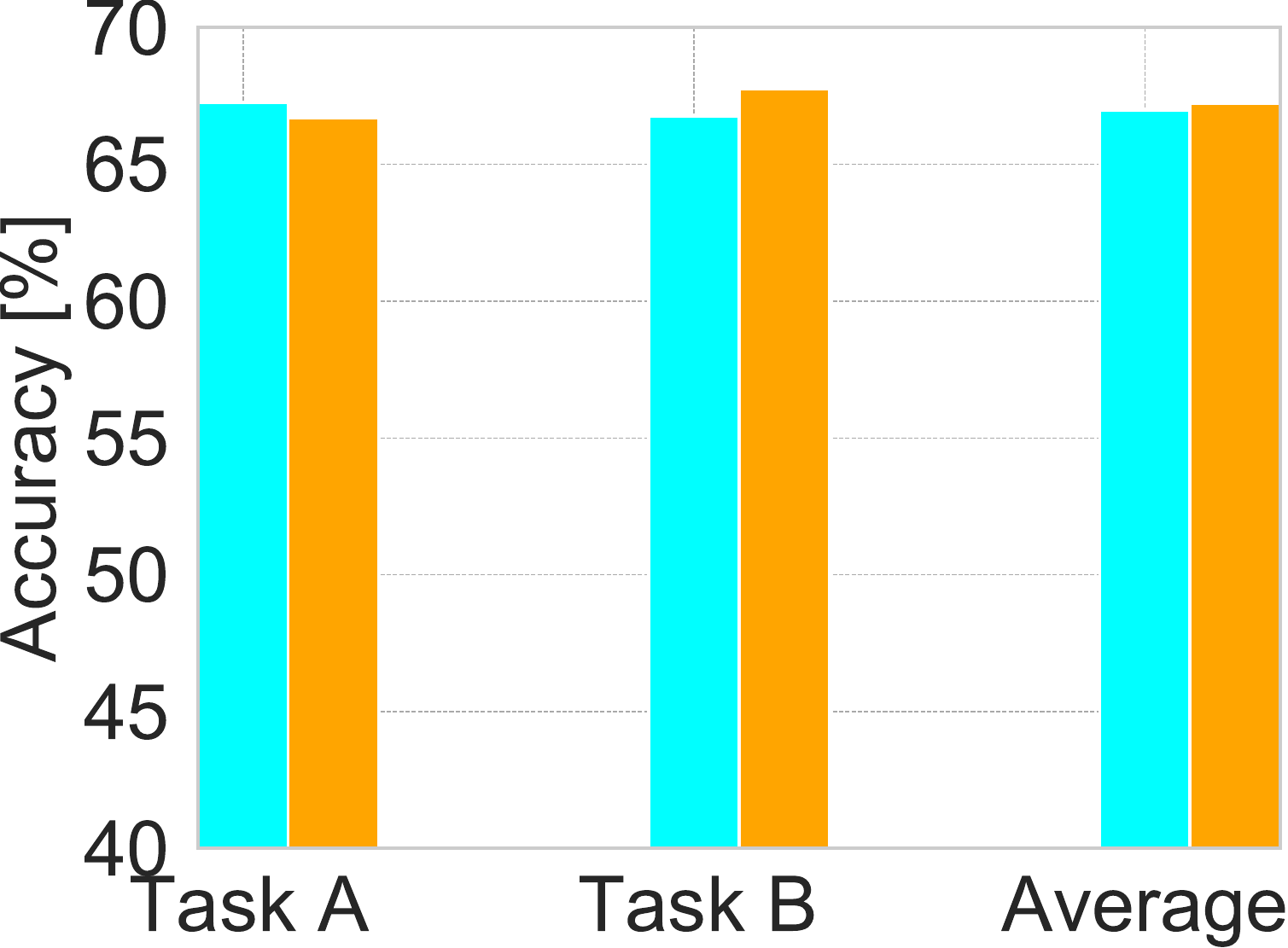}\hspace{0.0 cm}
    \includegraphics[width=0.23\linewidth]{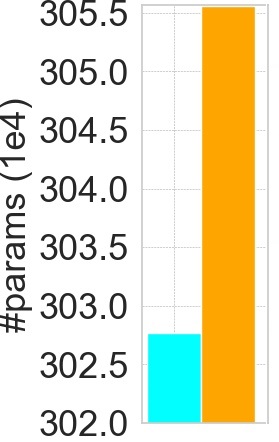}\hspace{0.0 cm}
    \includegraphics[width=0.21\linewidth]{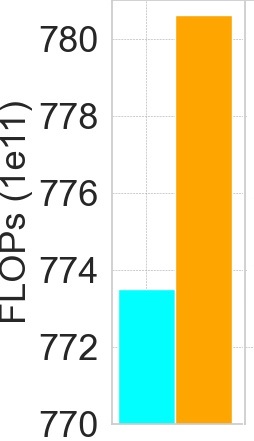}
    \centering
    \includegraphics[width=0.9\linewidth]{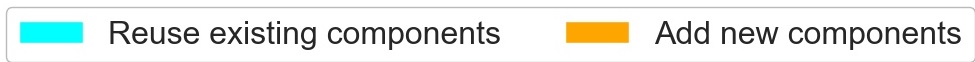}
    \caption{} 
    \label{fig:add_reuse}
  \end{subfigure}
   
  \caption{(a) Two different sub-networks are evaluated for Task B in class-IL. One sub-network balances forgetting and forward transfer, while the other maintains the performance of Task A at the expense of Task B. (b) Performing the same model-altering scheme in task-IL and class-IL leads to different performance. (c) By reusing some of Task A's components during learning Task B, we can achieve similar performance in class-IL as adding new components while reducing the memory and computational costs, represented by model parameters (\#params) and floating-point operations (FLOPs), respectively. Details are in Appendix \ref{sec:illustrative_experiments}.}
  \label{fig:illustrative_experiments}
\end{figure*}

Continual learning (CL) aims to build intelligent agents based on deep neural networks that can learn a sequence of
tasks. The main challenge in this paradigm is the stability-plasticity dilemma~\cite{mermillod2013stability}. Optimizing all model weights on a new task (\textit{highest plasticity}) causes forgetting of past tasks~\cite{mccloskey1989catastrophic}. While fixing all weights (\textit{highest stability}) hinders learning new tasks. Finding the balance between stability and plasticity is challenging. The challenge becomes more difficult when other CL requirements are considered, such as using fixed-capacity models without access to past data and limiting memory and computational costs \cite{hadsell2020embracing}. 

Task-specific components strategy~\cite{golkar2019continual,mallya2018packnet,sokar2021spacenet} offers some flexibility to address this dilemma by using different components (connections/neurons), i.e., sub-network within a model, for each task. The components of a new task are flexible to learn, while the components of past tasks are fixed. There are some challenges that need to be tackled in this strategy to balance multiple CL desiderata: 

\textbf{(a) Selection of a new sub-network.} Current methods focus solely on forgetting and choose a sub-network for a new task that would maintain the performance of past tasks regardless of its effectiveness for learning the new task. This hinders the forward transfer of relevant past knowledge in future learning.

\textbf{(b) Managing the fixed-capacity and training efficiency.} Typically, new components are added for every new task in each layer, which may unnecessarily consume the available capacity and increase the computational costs. Using fixed-capacity models for CL requires utilizing the capacity efficiently. 

\textbf{(c) Operating in the class-Incremental Learning setting (class-IL).} Unlike task-Incremental Learning (task-IL), where each task has a separate classifier under the assumption of the availability of task labels at inference, in class-IL, a \textit{unified} classifier is used for all classes in seen tasks so far. The latter is more realistic, yet it brings additional challenges due to agnosticity to task labels: \textbf{(1) class ambiguities}, using past knowledge in learning new classes causes ambiguities between old and new classes. \textbf{(2) component-agnostic inference}, all model components are used at inference since the task label is not available to select the corresponding components. This increases interference between tasks and the performance dependency on the sub-network of each task. 

In response to these challenges, we study the following question: \textit{How to alter the model structure when a new task arrives to balance CL desiderata in class-IL? Specifically, which components should be added, updated, fixed, or reused?} 

We summarize our findings below with illustrations shown in Figure \ref{fig:illustrative_experiments}:
\begin{itemize}
    \item The chosen sub-network of each task has a crucial role in the performance, affecting both forgetting and forward transfer (Figure 1a).
    \item The optimal altering of a model differs in task-IL and class-IL. For example, all components from past similar tasks could be \textit{reused} in learning a new task with minimal memory and computational costs in task-IL. However, this altering limits learning in class-IL due to class ambiguities (Figure \ref{fig:acc_TIL_CIL}). 
    \item Reusability of past \textit{relevant} components is applicable in class-IL under careful considerations for class ambiguity. It enhances memory and computational efficiency while maintaining performance (Figure \ref{fig:add_reuse}).
    \item Neuron-level altering is crucial in class-IL (e.g., \textit{fixing} some neurons for a task), while connection-level altering could be sufficient in task-IL since only one sub-network is selected at inference using the task label (Appendix \ref{appendix:connection_versus_neuron}).
    \item The challenge in balancing CL desiderata increases in class-IL when similarity across tasks increases (Section \ref{sec:results}). 
\end{itemize}

Motivated by these findings, we propose a new CL method, named AFAF, based on sparse sub-networks within a fix-capacity model to address the above-mentioned challenges. Without access to past data, AFAF aims to Avoid Forgetting and Allow Forward transfer in class-IL. In particular, when a new task arrives, we identify the relevant knowledge from past tasks and allocate a new sub-network that enables selective forward transfer of this knowledge while maintaining past knowledge. Moreover, we reuse some of the allocated components from past tasks. To enable selective forward transfer and component re-usability jointly with forgetting avoidance in class-IL, AFAF considers the extra challenges of class-IL (class-ambiguities and agnostic component inference) in model altering. We propose two variants of standard CL benchmarks to study the challenging case where high similarity exists across tasks in class-IL. Experimental results show that AFAF outperforms baseline methods on various benchmarks while reducing memory and computation costs. In addition, our analyses reveal the challenges of class-IL over task-IL that necessitate different model altering. 
 
\section{Related Work}
\label{sec:related_work}
We divide CL methods into two categories: replay-free and replay-based.

\textbf{Replay-free} In this category, past data is inaccessible during future learning. It includes two strategies:
\textbf{(1)} \textit{\textbf{Task-specific components.}} \textit{Specific} components are assigned to each task. Current methods either \textit{extend} the initially allocated capacity of a model for new tasks~\cite{rusu2016progressive,yoon2018lifelong} or use a \textit{fixed-capacity} model and add a \textit{sparse} sub-network for each task~\cite{mallya2018packnet,mallya2018piggyback,yoon2019scalable,mazumder2021few,wortsman2020supermasks,sokar2021spacenet}. Typically, connections of past tasks are kept \textit{fixed}, and the newly added ones are \textit{updated} to learn the current task. The criteria used for \textit{adding} new connections often focus solely on avoiding forgetting, limiting the selective transfer of relevant knowledge to learn future tasks. Moreover, most methods rely on task labels to pick the connections corresponding to the task at inference. SpaceNet~\cite{sokar2021spacenet} addressed component-agnostic inference by learning sparse representation during training and \textit{fixing} a fraction of the most important neurons of each task to reduce interference. Sparse representations are learned by training each sparse sub-network using dynamic sparse training~\cite{mocanu2018scalable,hoefler2021sparsity}, where weights and the sparse topology are jointly optimized. Yet, selective transfer and class ambiguity are not addressed. 
\textbf{(2)} \textit{\textbf{Regularization-based.}} A \textit{fixed-capacity} model is used, and \textit{all} weights are updated for each task. Forgetting is addressed by constraining changes in the important weights of past tasks~\cite{zenke2017continual,kirkpatrick2017overcoming,aljundi2018memory} or via distillation loss~\cite{li2017learning,dhar2019learning}. 

\textbf{Replay-based} In this category, forgetting is addressed by \textit{replaying}: (1) a subset of old samples~\cite{rebuffi2017icarl,lopez2017gradient,riemer2018learning,chaudhry2018efficient,bang2021rainbow}, (2) pseudo-samples from a generative model~\cite{mocanu2016online,shin2017continual,sokar2021learning}, or (3) generative high-level features~\cite{van2020brain}. A buffer is used to store old samples or a generative model to generate them.

Attention to task similarities and forward transfer has recently increased. Most efforts are devoted to task-IL. In \cite{ramasesh2020anatomy}, the analysis showed that higher layers are more prone to forgetting, and intermediate semantic similarity across tasks leads to maximal forgetting. SAM~\cite{sokar2021self} meta-trains a self-attention mechanism for selective transfer in dense networks. CAT~\cite{ke2020continual} addresses the relation between task similarities and forward/backward transfer in task-IL, where task labels are used to find similar knowledge in dense models. In \cite{lee2021sharing}, an expectation-maximization method was proposed to select the shared or added \textit{layers} to promote transfer in task-IL. Similarly, \cite{veniat2021efficient} uses a modular neural network architecture and searches for the optimal path for a new task by the composition of neural modules. A task-driven method is used to reduce the exponential search space. In~\cite{chen2020long}, the lottery ticket hypothesis~\cite{frankle2018lottery} is studied for CL. 

\section{Network Structure Altering}
\label{sec:network_altering}
When a model faces a new task, task-specific components methods alter its structure via some actions to address the stability-plasticity dilemma. Next, we will present commonly used and our proposed actions. 
\subsection{Connection-Level Actions}
\begin{figure*}[h]
  \begin{subfigure}[b]{0.2\columnwidth}
    \includegraphics[width=0.9\linewidth]{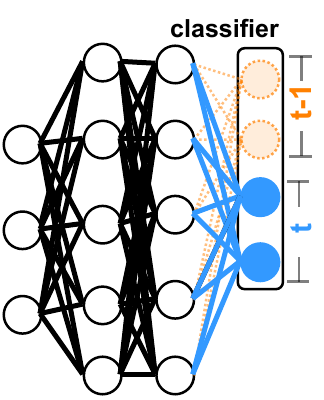}
    \caption{\scriptsize{Dense}}
    \label{fig:replay_free_reg}
  \end{subfigure}
  \hfill 
  \begin{subfigure}[b]{0.2\columnwidth}
    \includegraphics[width=0.9\linewidth]{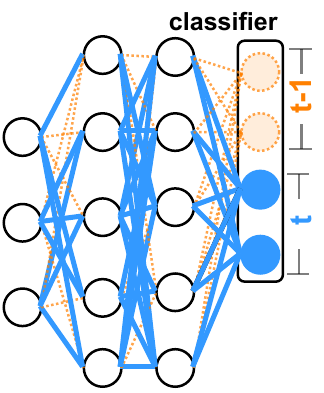}
    \caption{\scriptsize{Dense+Prune}}
    \label{fig:replay_free_packNet}
  \end{subfigure}
    \hfill
    \begin{subfigure}[b]{0.2\columnwidth}
    \includegraphics[width=0.9\linewidth]{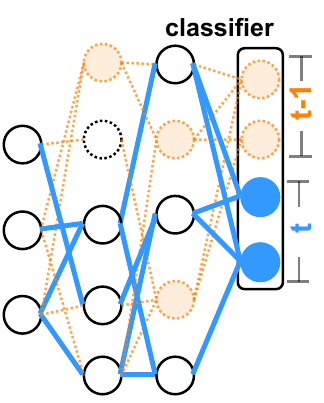}
    \caption{\scriptsize{Sparse}}
    \label{fig:replay_free_spaceNet}
  \end{subfigure}
   \hfill
    \begin{subfigure}[b]{0.25\columnwidth}
    \includegraphics[width=0.95\linewidth]{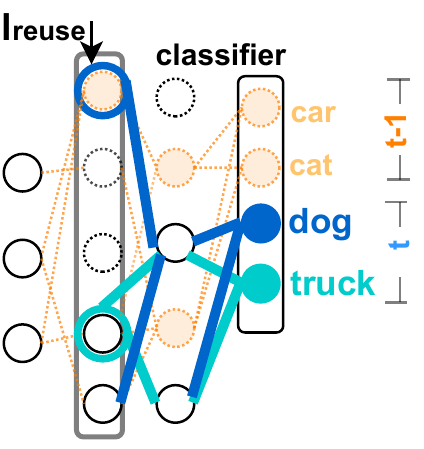}
    \caption{\scriptsize{Sparse+Reuse}}
    \label{fig:replay_free_AFAF}
    \end{subfigure}
  \hfill
      \begin{subfigure}[b]{\columnwidth}
      \centering
    \includegraphics[width=0.8\linewidth]{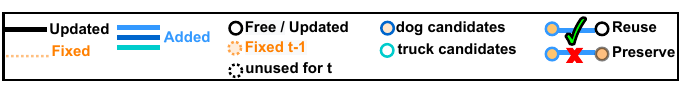}
    \label{fig:replay_free_legend}
  \end{subfigure}
  \caption{Overview of a model when it faces a new task $t$ using different methods. (a) Regularization methods use and update all connections with regularization for learning task $t$ \cite{kirkpatrick2017overcoming,aljundi2018memory}. (b) Dense connections are added, and unimportant connections are pruned after training \cite{mallya2018packnet}. Connections of previous tasks are fixed. (c) Sparse connections are added from scratch \cite{sokar2021spacenet}.  Connections and important neurons of past tasks are fixed (dashed lines and filled circles). (d) Our proposed method, AFAF. Starting from layer $l_{reuse}$, new connections are added for each class using the candidate neurons for the class to enable selective transfer and free neurons to capture residual knowledge. New outgoing connections are allowed from fixed neurons at layer $l$ but could only connect free neurons (unfilled circles) in layer $l+1$ to preserve the old knowledge.}
  \label{fig:diff_replay_free_strategy}
\end{figure*}

Most state-of-the-art methods alter fixed-capacity models at the \textbf{\textit{connection level}}. The connection-level actions include:
\begin{itemize}
\item\enquote{\textbf{Add}}: New connections, parameterized by $\textbf{W}^t$, are added for each new task $t$. Each task has a \textit{sparse} sub-network resulting from either pruning dense connections \cite{mallya2018packnet} (Figure \ref{fig:replay_free_packNet}) or adding sparse ones from scratch \cite{sokar2021spacenet} (Figure \ref{fig:replay_free_spaceNet}). The current practice is to \textbf{\textit{randomly}} add connections in \textit{each} layer using unimportant components of past tasks focusing solely on forgetting \cite{sokar2021spacenet,mallya2018packnet}; $\textbf{W}^t = \{\textbf{W}_l^t:1 \leq l \leq L-1\}$ where $L$ is the number of layers. This will be addressed in Section \ref{sec:method} to enable \textbf{\textit{selective transfer}} and \textbf{\textit{reusability}}.
\item\enquote{\textbf{Update}}: $\textbf{W}^t$ are updated during task $t$ training (i.e., allows plasticity). 
\item\enquote{\textbf{Fix}}: Once a task has been learned, $\textbf{W}^t$ are frozen (i.e., controls stability). 
\end{itemize}
Note that regularization methods \textit{add} \textbf{dense} connections at step $t=0$ (Figure \ref{fig:replay_free_reg}). Each task \textit{updates} all weights. Stability is controlled via regularization. 

\subsection{Neuron-Level Actions}
Component-agnostic inference in class-IL makes reducing task interference at the connection level only more challenging since connections from different tasks share the same neurons (Figure \ref{fig:replay_free_packNet}). Reducing interference at the \textit{representational} level is more efficient (Appendix \ref{appendix:connection_versus_neuron}). Hence, we believe that the following neuron-level actions are needed:
\begin{itemize}
\item\enquote{\textbf{Fix}}: After learning a task, its \textit{important} neurons should be frozen to reduce the drift in its representation \cite{sokar2021spacenet}. Other neurons are \enquote{free} to be updated. 
\item\enquote{\textbf{Reuse}}: Reusing neurons that capture useful representation in learning future tasks. Fixed neurons could be reused by allowing \textbf{\textit{outgoing}} connections only from these neurons. Details are provided in the next section.    
\end{itemize}

Most previous methods operate at the connection level except the recent work, SpaceNet \cite{sokar2021spacenet}, which shows its favorable performance over replay-free methods using a fixed-capacity model via \textit{fixing} the important neurons of each task and learning sparse representations (Figure \ref{fig:replay_free_spaceNet}). However, besides adding new components for every task in each layer, it does not \textit{selectively} transfer relevant knowledge. A summary is provided in Table \ref{table:actions}. Our goal is to address selective transfer and reusability of previous relevant neurons while maintaining them stable.
\begin{table*}[t]
\centering
\caption{Actions used to alter the model components at the connection and neuron levels by different methods.}
\label{table:actions}
\resizebox{\columnwidth}{!}{

\begin{tabular}{|l|c|c|c|c|c|}
\hline

Method & Add weights & Fix old weights & Fix neurons & Reuse past components & Selective transfer \\
\hline
EWC \cite{kirkpatrick2017overcoming}, MAS \cite{aljundi2018memory}  & Dense & $\times$ &  $\times$ &   $\times$ & $\times$  \\\hline
PackNet \cite{mallya2018packnet} & Dense+Prune & $\surd$ & $\times$ & $\times$ & $\times$ \\\hline
SpaceNet \cite{sokar2021spacenet} & Sparse & $\surd$ & $\surd$ & $\times$ & $\times$   \\\hline
AFAF (ours) & Sparse & $\surd$ & $\surd$ & $\surd$& $\surd$ \\
\hline
\end{tabular}
}
\end{table*}
\section{Proposed Method}
\label{sec:method}
We consider the problem of learning $T$ tasks sequentially using a \textit{fix-capacity} model with a unified classifier. Each task $t$ brings new $C$ classes. The output layer is extended with new $C$ neurons. Let $\mathcal{D}^t = \{D^{t_c}\}|_{c=1}^C$, where $D^{t_c}$ is the data of class $c$ in task $t$. Once a task has been trained, its data is \textit{discarded}. 

AFAF is a new task-specific component method that dynamically trains a sparse sub-network for each task from scratch. Our goal is to alter the model components to (i) reuse past components and (ii) selectively transfer relevant knowledge while (iii) reducing class ambiguity in class-IL. This relies on the selection of a new sub-network and its training. When a new task arrives, AFAF adds a sub-network such that past knowledge is transferred from relevant neurons while retaining it, and residual knowledge is learned. Section \ref{sec:adding_new_componenets} introduces the selection mechanism of a sub-network and illustrates the added and reused components. Section \ref{sec:class_ambig} discusses the considerations for class ambiguity. In Section \ref{sec:training}, we present task training and identify the fixed and updated components.    

\subsection{Selection of a New Sub-network}
\label{sec:adding_new_componenets}
\textbf{Notation} Let $l \in \{1,..,L\}$ represents a layer in a neural network model. A sparse connections $W_l^t$, with a sparsity $\epsilon_l$, are allocated for task $t$ in layer $l$ between a subset of the neurons denoted by $\textbf{h}_l^{alloc}$ and $\textbf{h}^{alloc}_{l+1}$ in layers $l$ and $l+1$, respectively. We use the term \textit{neuron} to denote a node in multilayer perceptron networks or a feature map in convolution neural networks (CNNs). The selected neurons in each layer determine the initial sub-network for a new task. 

Typically, when there is a semantic similarity between old and new classes, the existing learned components likely capture some useful knowledge for the current task. Hence, we propose to reuse some of these components. Namely, instead of adding new connections in each layer, we allocate new sparse connections starting from layer $l_{reuse}$ (Figure \ref{fig:replay_free_AFAF}). $l_{reuse}$ is a hyper-parameter that controls the trade-off between adding new connections and reusing old components based on existing knowledge and the available capacity. Hence, the connections of a new task $t$ are $\textbf{W}^t = \{\textbf{W}_l^t:l_{reuse} \leq l \leq L-1\}$. Layers from $l=1$ up to but excluding layer $l_{reuse}$ remains unchanged. We show that reusing past components reduces memory and computational costs (Section \ref{sec:results}), balances forward and backward transfer (Section \ref{sec:analysis}), and utilizes the available fixed-capacity efficiently by allowing new dissimilar tasks to acquire more resources that are saved by reusability, i.e., higher density for sparse sub-networks (Appendix \ref{sec:utilize_model}).

Starting from $l_{reuse}$, we add new sparse connections $\textbf{W}_l^t$ in each layer $l$. The selected neurons for allocation should allow: (i) selective transfer of relevant knowledge to promote forward transfer and (ii) preserving old representation to avoid forgetting. To this end, we select a set of \textit{candidate} and \textit{free} neurons in each layer, as we will discuss next. Details are provided in Algorithm \ref{AFAF_connection_allocation_alg}.

\textbf{Identify Candidate Neurons.} To \textit{selectively transfer} the relevant knowledge, for each layer $l \geq l_{reuse}$, we identify a set of candidate neurons $\mathcal{R}_{l}^c$ that has a high potential of being useful when \enquote{reused} in learning \textit{class} $c$ in a new task $t$. Classes with semantic similarities are most likely to share similar representations (Appendix \ref{appendix:hiddenrepresentation}). Hence, we consider the average activation of a neuron as a metric to identify its potential for reusability. In particular, we feed the data of each class $c$ in a new task $t$, $\mathcal{D}^{t_c}$, to the trained model at time step $t-1$, $f^{t-1}(\textbf{W}^{t-1})$, and calculate the average activation $\mathcal{A}^c_l$ in each layer $l$ as follows:
\begin{equation}
  \mathcal{A}_{{l}}^c(\textbf{W}^{t-1}) =  \frac{1}{|\mathcal{D}^{t_{c}}|} \sum_{i=1}^{|\mathcal{D}^{t_{c}}|} \textbf{a}_{l}(x^{t_{c}}_i),
  \label{eq:activation_cand}
\end{equation}
where $\textbf{a}_{l}$ is the vector of neurons activation at layer $l$ when a sample $x^{t_{c}}_i \in \mathcal{D}^{t_{c}}$ is fed to the model $f^{t-1}(\textbf{W}^{t-1})$, and $|\mathcal{D}^{t_{c}}|$ is the number of samples of class $c$. Once the activation is computed, neurons with the top-$\kappa$ activation are selected as potential candidates $\mathcal{R}_{l}^c$ as follows:
\begin{equation}
  \mathcal{R}_{l}^c = \{i|\mathcal{A}_{l_i}^c\in\textnormal{top-$\kappa$}(\mathcal{A}_{l}^c)\},
  \label{eq:cand_neurons}
\end{equation}
where $\kappa$ denotes the number of candidate neurons, and $\mathcal{A}_{l_i}^c$ is the average activation of neuron $i$ in layer $l$. Neuron activity shows its effectiveness as an estimate of a neuron/connection importance in pruning neural networks for single task learning \cite{hu2016network,luo2017thinet,dekhovich2021neural}. To assess our choice of this metric to identify the candidate neurons, we compare against two metrics: \textbf{(1) Random}, where the candidate neurons are randomly chosen from all neurons in a layer, and \textbf{(2) Lowest}, where the candidates are the neurons with the lowest activation (See Section \ref{sec:analysis}). 

Note that by exploiting the representation of candidate neurons $\mathcal{R}_{l_{reuse}}^{c}$ in layer $l_{reuse}$, we selectively \textit{reuse} past components connected to these neurons in preceding layers ($l < l_{l_{reuse}}$). Hence, these components are stable but reusable. 

\textbf{Identify Free Neurons.} To preserve past representation while allowing reusability of neurons, \enquote{fixed} neurons that might be selected as candidates should be stable. Hence, we allow \textit{outgoing} connections from these neurons but not incoming connections. The outgoing connections from fixed neurons in layer $l$ can be connected to \enquote{free} neurons in layer $l+1$. To this end, in each layer, we select a subset of the free neurons for \textit{each task}, denoted as $\mathcal{S}_{l}^{Free}$. For these neurons, incoming and outgoing connections are allowed to be allocated and used to capture the specific representation of a new task.

\textbf{Allocation.} The neurons used to allocate new sparse weights $\textbf{W}^c_l$ between layers $l$ and $l+1$ for a class $c$ are as follows: 
\begin{algorithm}[h]
\scriptsize
\caption{AFAF Sub-network Allocation}
\label{AFAF_connection_allocation_alg}
\begin{algorithmic}[1]
\State \textbf{Require:} $l_{reuse}$, sparsity level $\epsilon_l$, $|\textbf{h}^{alloc}_l|$, $\kappa$, $\textbf{h}_{l}^{Fix}$  %
\ForEach{class $c$ in task $t$} \algorithmiccomment{Get candidate neurons}
\State $\mathcal{A}_{l}^c \gets$ calculate average activation of $\mathcal{D}^{t_c}$ \algorithmiccomment{Eq \ref{eq:activation_cand}}
\State $\mathcal{R}_{l}^c \gets $ get candidates for class $c$ $\forall l \geq l_{reuse}$ \algorithmiccomment{Eq \ref{eq:cand_neurons}}
\EndForEach
\State $\mathcal{S}_{l}^{Free} \gets$ randomly select subset of free neurons $\forall l \geq l_{reuse}$
\State $\mathcal{R}_{L-1}^c, \mathcal{R}_{L}^c \gets \emptyset$ \algorithmiccomment{No candidate neurons used in last layer}
\State $\mathcal{S}_{L}^{Free} \gets \{c\}$ $\quad$ $\forall c\in t$ \algorithmiccomment{Output neurons for new classes}
\ForEach{class $c$ in task $t$}
\For{$l \gets l_{reuse}$ to $L-1$}
\State $\textbf{h}^{alloc}_{l} \gets$ \{$\mathcal{R}_{l}^c \cup	\mathcal{S}_{l}^{Free} $\} \algorithmiccomment{Neurons for allocation}
\State $\textbf{h}^{alloc}_{l+1} \gets$ $\{(\mathcal{R}_{l+1}^{c}\setminus \textbf{h}_{l+1}^{Fix}) \cup \mathcal{S}_{l+1}^{Free} \}$ \algorithmiccomment{Neurons for allocation}
\State Allocate $\textbf{W}^{c}_{l}$ with sparsity $\epsilon_l$ between $\textbf{h}^{alloc}_{l}$ \& $\textbf{h}^{alloc}_{l+1}$
\State $\textbf{W}_{l} \gets \textbf{W}_{l} \cup \textbf{W}^{c}_{l} $  
\EndForEach
\EndForEach
\end{algorithmic}
\end{algorithm}

\begin{equation}
 \begin{gathered}
 \begin{aligned}
   \textbf{h}_{l}^{alloc}&=\{\mathcal{R}_{l}^{c} \cup \mathcal{S}_{l}^{Free}\},  \\ 
   \textbf{h}_{l+1}^{alloc} &= \{(\mathcal{R}_{l+1}^{c}\setminus \textbf{h}_{l+1}^{Fix}) \cup \mathcal{S}_{l+1}^{Free} \}, 
   \end{aligned}
   \end{gathered}
  \label{eq:alloc}
\end{equation}
where $\textbf{h}_{l+1}^{Fix}$ is the set of fixed neurons at layer $l+1$.
\subsection{Addressing Class Ambiguities}
\label{sec:class_ambig}
Unlike task-IL, reusing past relevant knowledge in future learning may result in class ambiguity in class-IL (Section \ref{sec:intro}). To this end, we propose three constraints for reusability that aim to (1) allow a new task to learn its specific representation and (2) increase the decision margin between classes (see Section \ref{sec:class_ambiguities} for analysis). 

\textbf{Learn specific representation.} Learning specific representation reduces ambiguity between similar classes. Hence, we add two constraints. First, new connections should be added in at least one layer before the classification layer to capture the specific representation of the task (i.e., $l_{reuse}\in [2,L-2]$). Second, the output connections are allocated using free neurons only (i.e., no candidate neurons are used; $\textbf{h}^{alloc}_{L-1}=\mathcal{S}_{L-1}^{Free}$) since candidate neurons in the highest-level layer are highly likely to capture the specific representation of past classes. 

\textbf{Increase the decision margin between classes.} To learn more discriminative features and increase the decision margin between classes, we use orthogonal weights in the output layer \cite{li2020oslnet}. To this end, once a task has been trained, we force all its neurons in the last layer, $\textbf{h}^{alloc}_{L-1}$, to be fixed and not reusable.
\subsection{Training}
\label{sec:training}
The new connections are trained with stochastic gradient descent. During training, the weights and important neurons of past tasks are kept fixed to protect past representation. We follow the approach in \cite{sokar2021spacenet} to train a sparse topology (connections distribution) and identify the portion of the important neurons from $\textbf{h}_l^{alloc}$ in each layer that will be fixed after training (Appendix \ref{appendix:SpaceNet_DST_neuronImp}). In short, a sparse topology is optimized by a dynamic sparse training approach to produce sparse representation. To reuse important neurons of past tasks while protecting the representation, we block the gradient flow through all-important neurons of past tasks, even if they are reused as candidates for the current task. The gradient $\textbf{g}_{l}$ through the neurons of layer $l$ is: 
\begin{equation}
\textbf{g}_{l} = \textbf{g}_l  \otimes  (1-\mathbbm{1}_{\textbf{h}_{l}^{Fix}}),
\end{equation}
where $\mathbbm{1}$ is the indicator function. This \textit{allows} not to forget past knowledge while reusing it for selective transfer. Since the important neurons of dissimilar classes are less likely to be involved in the sub-network selection, they are protected.
\section{Experiments and Results}
\textbf{Baselines.} Our study focuses on the replay-free setting using a fixed-capacity model. Therefore, we compare with several representative regularization methods that use \textit{dense} fixed-capacity, \textbf{EWC} \cite{kirkpatrick2017overcoming}, \textbf{MAS} \cite{aljundi2018memory}, and \textbf{LWF} \cite{li2017learning}. In addition, we compare with task-specific components methods that use \textit{sparse} sub-networks within a fixed-capacity model, \textbf{PackNet} \cite{mallya2018packnet} and \textbf{SpaceNet} \cite{sokar2021spacenet}.     

\textbf{Benchmarks.} We performed our experiments on three sets of benchmarks: (1) standard split-CIFAR10, (2) sequences with high semantic similarity at the class level across tasks, and (3) sequence of mixed datasets where tasks come from different domains to study the stability-plasticity dilemma for sequences with concept drift and interfering tasks.

\textbf{Standard Evaluation.} We evaluate the standard \textbf{split-CIFAR10} benchmark with 5 tasks. Each task consists of 2 consecutive classes of CIFAR10 \cite{krizhevsky2012imagenet}.

\textbf{Similar Sequences.} To assess replay-free methods under more challenging conditions, we design two new benchmarks with high semantic similarity across tasks. In the absence of past data, we test the unified classifier's ability to distinguish between similar classes when they are not presented together within the same task. \textbf{(1) sim-CIFAR10} is constructed from CIFAR-10 by shuffling the order of classes to increase the similarity across tasks (Appendix \ref{appendix:experimental_setup}; Table \ref{table:details_sim_CIFAR10}). It consists of 5 tasks. \textbf{(2) sim-CIFAR100} is constructed from CIFAR-100 \cite{krizhevsky2009learning}. Classes within the same superclass in CIFAR-100 have high semantic similarity. Hence, we construct a sequence of 8 tasks with two classes each and distribute the classes from the same superclass in different tasks (Appendix \ref{appendix:experimental_setup}; Table \ref{table:details_sim-CIFAR100}). 

\textbf{Mix datasets.} The considered datasets are CIFAR10 \cite{krizhevsky2009learning}, MNIST \cite{lecun1998mnist}, NotMNIST \cite{notMNIST}, and FashionMNIST \cite{xiao2017fashion}. We construct a sequence of 8 tasks with 5 classes each. The first four tasks are dissimilar, while the second four are similar to the first ones (Appendix \ref{appendix:experimental_setup}; Table \ref{table:details_mix}). 

\textbf{Implementation Details.} Motivated by the recent study on architectures for CL \cite{mirzadeh2022architecture}, we follow \cite{serra2018overcoming,ke2020continual,veniat2021efficient} to use an AlexNet-like architecture \cite{krizhevsky2012imagenet} that is trained from scratch using stochastic gradient descent. We start reusing relevant knowledge in future tasks after learning similar ones. Therefore, for Mix and sim-CIFAR100, we start reusing past components from task 5, while for split-CIFAR10 and sim-CIFAR10, we start from task 3. Earlier tasks add connections in each layer using the free neurons. For all benchmarks, we use $l_{reuse}$ of 4. 

\textbf{Evaluation Metrics.} To evaluate different CL requirements, we assess various metrics: (1) \textbf{Average accuracy (ACC)}, which measures the performance at the end of the learning experience, (2) \textbf{Backward Transfer} \textbf{(BWT)} \cite{lopez2017gradient}, which measures the influence of learning a new task on previous tasks (large negative BWT means forgetting), (3) \textbf{Floating-point operations (FLOPs)}, which measure the required computational cost, and (4) \textbf{Model size (\#params)}, which is the number of model parameters. More details are in Appendix \ref{appendix:experimental_setup}.
 
\subsection{Results}
\label{sec:results}
Table \ref{table:results} shows the performance on each benchmark. AFAF consistently outperforms regularization-based methods and other task-specific components methods on all benchmarks. The difference in performance between split-CIFAR10 and sim-CIFAR10, which have the same classes in a different order, reveals the challenge caused by having similar classes across tasks in class-IL. All studied methods have lower ACC and BWT on sim-CIFAR10 than split-CIFAR10. Yet, AFAF is the most robust method towards this challenge. When the similarity across tasks increases more, as in sim-CIFAR100, regularization methods and PackNet fail to achieve a good performance. We also observe that LWF outperforms other regularization-based methods in most cases except on the Mix datasets benchmark, where there is a big distribution shift across tasks from different domains. Most task-specific components methods outperform the regularization-based ones with much lower forgetting. 

Analyzing task-specific components methods, we observe that altering the model at the connection level only by PackNet is not efficient in class-IL despite its high performance in task-IL (Appendix \ref{appendix:connection_versus_neuron}). Besides the additional memory and computational overhead of pruning and retraining dense models, the performance is lower than other task-specific components methods. Altering at the connection and neuron levels, as in SpaceNet and AFAF, enables higher performance. The gap between these methods increases when the sequence has a larger number of tasks with high similarities (i.e., sim-CIFAR100 and Mix datasets). 

AFAF consistently achieves higher ACC and BWT than SpaceNet on all benchmarks with various difficulty levels. Interestingly, the achieved performance is obtained by \textit{reusing} relevant knowledge via selective transfer. AFAF exploits the similarity across tasks in learning future tasks while addressing class ambiguities. Moreover, the performance gain is accompanied by using a smaller memory and less computational cost than all the baselines. 
\begin{table*}[t]
\caption{Evaluation results on four CL benchmarks in the replay-free  class-IL setting with fixed-capacity models. }
\label{table:results}
\centering
\resizebox{\columnwidth}{!}{
\begin{tabular}{|l|ccccccc||ccccccc|}
\hline
  & \multicolumn{7}{c||}{Split-CIFAR10} &  \multicolumn{7}{c|}{Sim-CIFAR10}  \\
  \hline
Method & ACC ($\uparrow$)& & BWT ($\uparrow$) &  &  FLOPs ($\downarrow$)& & \#params ($\downarrow$) & ACC ($\uparrow$)& & BWT ($\uparrow$) &  &  FLOPs ($\downarrow$)& & \#params ($\downarrow$) \\
\hline
Dense Model & - & & - &  & 1$\times$(14.97e14) & & 1$\times$(23459520) & -& & - & &  1$\times$(14.97e14) & & 1$\times$(23459520)\\
\hline
EWC \cite{kirkpatrick2017overcoming} & 38.30$\pm$0.81 & & -59.30$\pm$2.03 & &  1$\times$ & & 1$\times$ & 28.90$\pm$3.11 & &-66.10$\pm$5.49 & &1$\times$ & & 1$\times$\\
LWF  \cite{li2017learning} &  48.10$\pm$2.28& & -42.33$\pm$1.15&  &   1$\times$ & & 1$\times$ & 40.43$\pm$1.22& &-50.36$\pm$1.98 & & 1 $\times$ & & 1$\times$\\
MAS \cite{aljundi2018memory} & 38.30$\pm$1.06& &-56.56$\pm$3.30  & &   1$\times$ & & 1$\times$ & 28.93$\pm$4.05& &-61.86$\pm$5.66 & &  1$\times$ & & 1$\times$\\
PackNet \cite{mallya2018packnet} & 44.33$\pm$0.85& & -50.40$\pm$1.45& &  3.081$\times$ & & 1$\times$ & 32.63$\pm$1.22& &-61.96$\pm$1.52 & & 3.081$\times$ & & 1$\times$\\
SpaceNet \cite{sokar2021spacenet} & 51.63$\pm$1.28& & -36.50$\pm$1.53 & &   0.154$\times$ & & 0.154$\times$ & 42.86$\pm$4.57 & & -30.69$\pm$4.63 & & 0.325$\times$ & & 0.325$\times$ \\
AFAF (ours)& \textbf{52.35$\pm$2.35}& & \textbf{-32.93$\pm$3.19} &  &  \textbf{0.148}$\times$ & & \textbf{0.148$\times$} & \textbf{45.23$\pm$2.14} & &\textbf{-29.35$\pm$3.54}& & \textbf{0.322$\times$} & & \textbf{0.322$\times$}\\
\hline
\end{tabular}
}

\resizebox{\columnwidth}{!}{
\centering
\begin{tabular}{|l|ccccccc||ccccccc|}
\hline
  & \multicolumn{7}{c||}{Sim-CIFAR100} & \multicolumn{7}{c|}{Mix datasets} \\
\hline
Method & ACC ($\uparrow$)& & BWT ($\uparrow$) &  &  FLOPs ($\downarrow$)& & \#params ($\downarrow$) & ACC ($\uparrow$)& & BWT ($\uparrow$) &  &  FLOPs ($\downarrow$)& & \#params ($\downarrow$) \\
\hline
Dense Model & -& & - & & 1$\times$(23.96e13)& & 1$\times$(23471808) & - & & - & & 1$\times$(5.600e15)&  & 1$\times$(23520960)  \\
\hline
EWC \cite{kirkpatrick2017overcoming} & 15.73$\pm$1.89& & -74.50$\pm$5.10 &  & 1$\times$ & &1$\times$&  54.63$\pm$1.93 & &-31.33$\pm$2.82&  & 1$\times$ & &1$\times$\\
LWF \cite{li2017learning} &  14.40$\pm$2.69& & -52.86$\pm$1.12 &  & 1$\times$ & &1$\times$&  40.60$\pm$3.25 & &-60.46$\pm$3.54&  & 1$\times$ & &1$\times$\\
MAS \cite{aljundi2018memory} &  13.50$\pm$0.66& & -81.70$\pm$1.02 &  & 1$\times$ & &1$\times$&  56.86$\pm$1.81 & &-26.83$\pm$3.44 & & 1$\times$ & &1$\times$ \\
PackNet \cite{mallya2018packnet}& 10.12$\pm$2.55& & -20.35$\pm$3.83 & & 5.241$\times$ & & 0.8$\times$&16.61$\pm$2.35& & \textbf{-18.58$\pm$1.38}&  & 5.250$\times$ & & 0.8$\times$\\
SpaceNet \cite{sokar2021spacenet}& 32.86$\pm$2.73& & -36.29$\pm$2.78 & & 0.089$\times$ & &0.089$\times$& 56.25$\pm$1.69& & -30.02$\pm$1.79&  & 0.053$\times$ & &0.053$\times$\\
AFAF (ours)& \textbf{33.74$\pm$2.18}&  & \textbf{-21.26$\pm$2.21} & &\textbf{0.088$\times$} & & \textbf{0.088$\times$}& \textbf{59.02$\pm$1.76} & &-25.91$\pm$1.51&  & \textbf{0.050$\times$} & & \textbf{0.050$\times$}\\
\hline
\end{tabular}
}
\end{table*}

\begin{figure}
\centering
  \begin{subfigure}[b]{0.8\columnwidth}
    \includegraphics[width=0.3\linewidth]{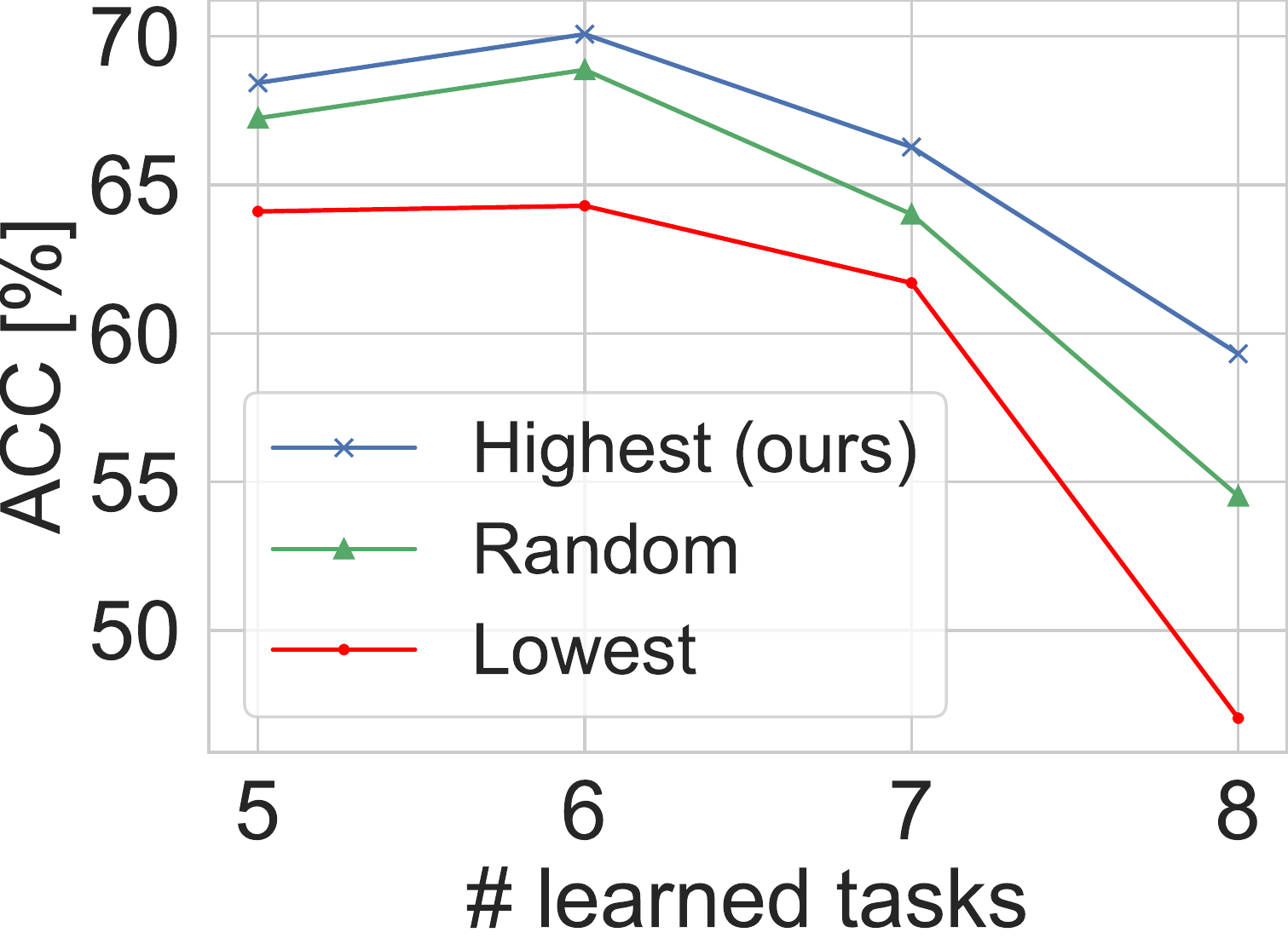} \hfill
        \includegraphics[width=0.3\linewidth]{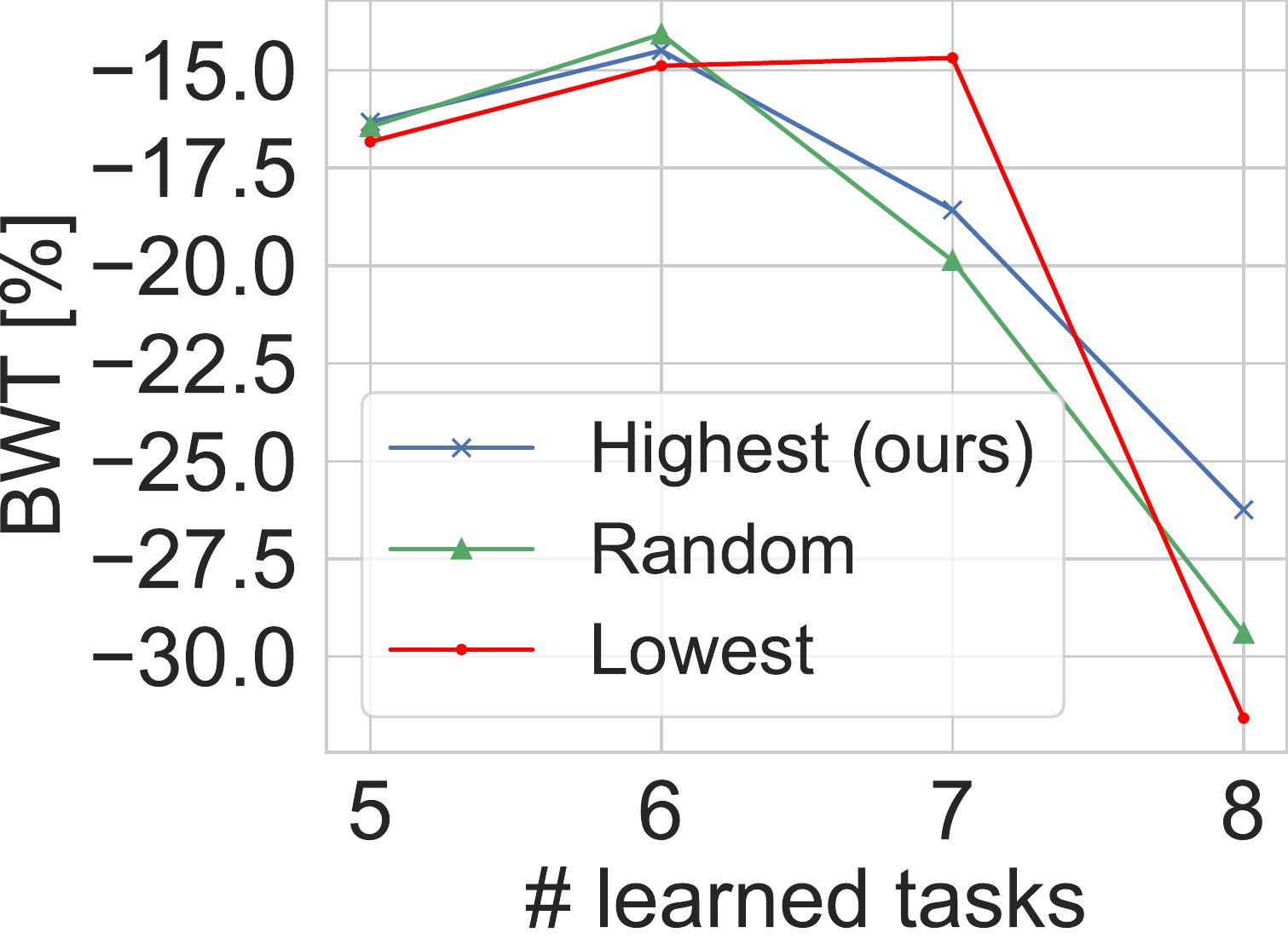} \hfill
         \includegraphics[width=0.3\linewidth]{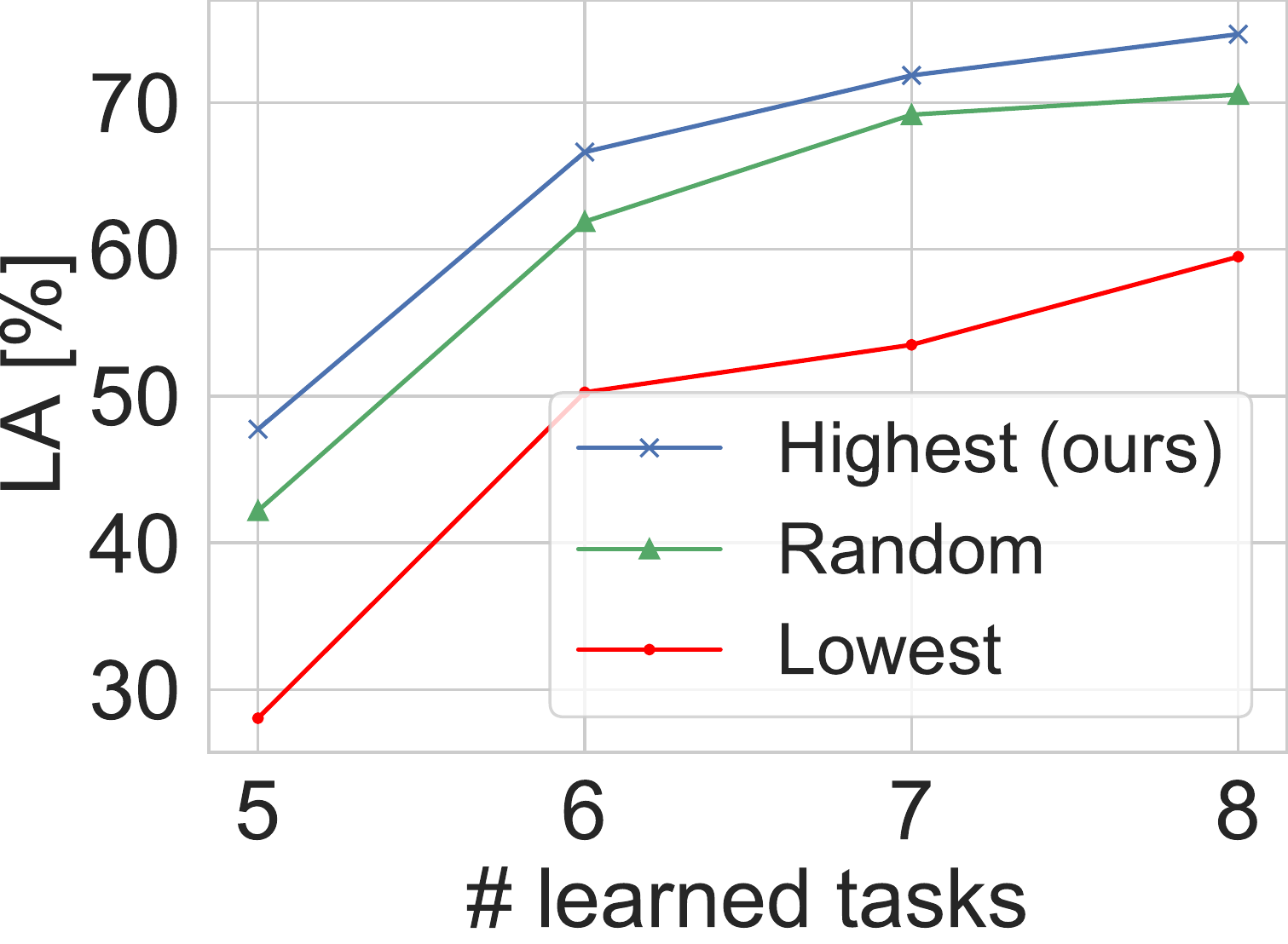} 

    \caption{Mix datasets}
    \label{fig:Mix_random_lowest}
  \end{subfigure}
  \hfill 
  \begin{subfigure}[b]{0.8\columnwidth}
    \includegraphics[width=0.3\linewidth]{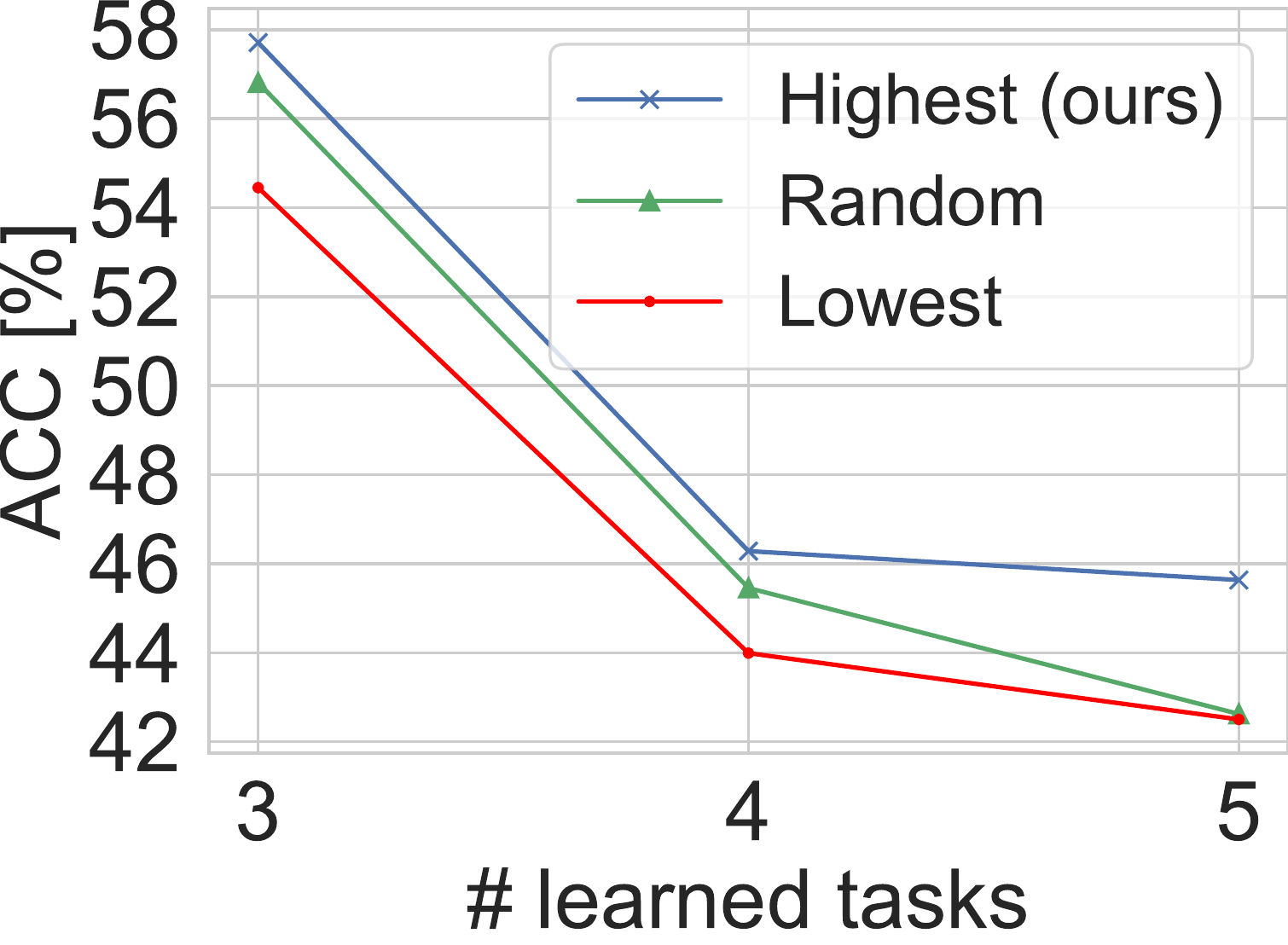} \hfill
    \includegraphics[width=0.3\linewidth]{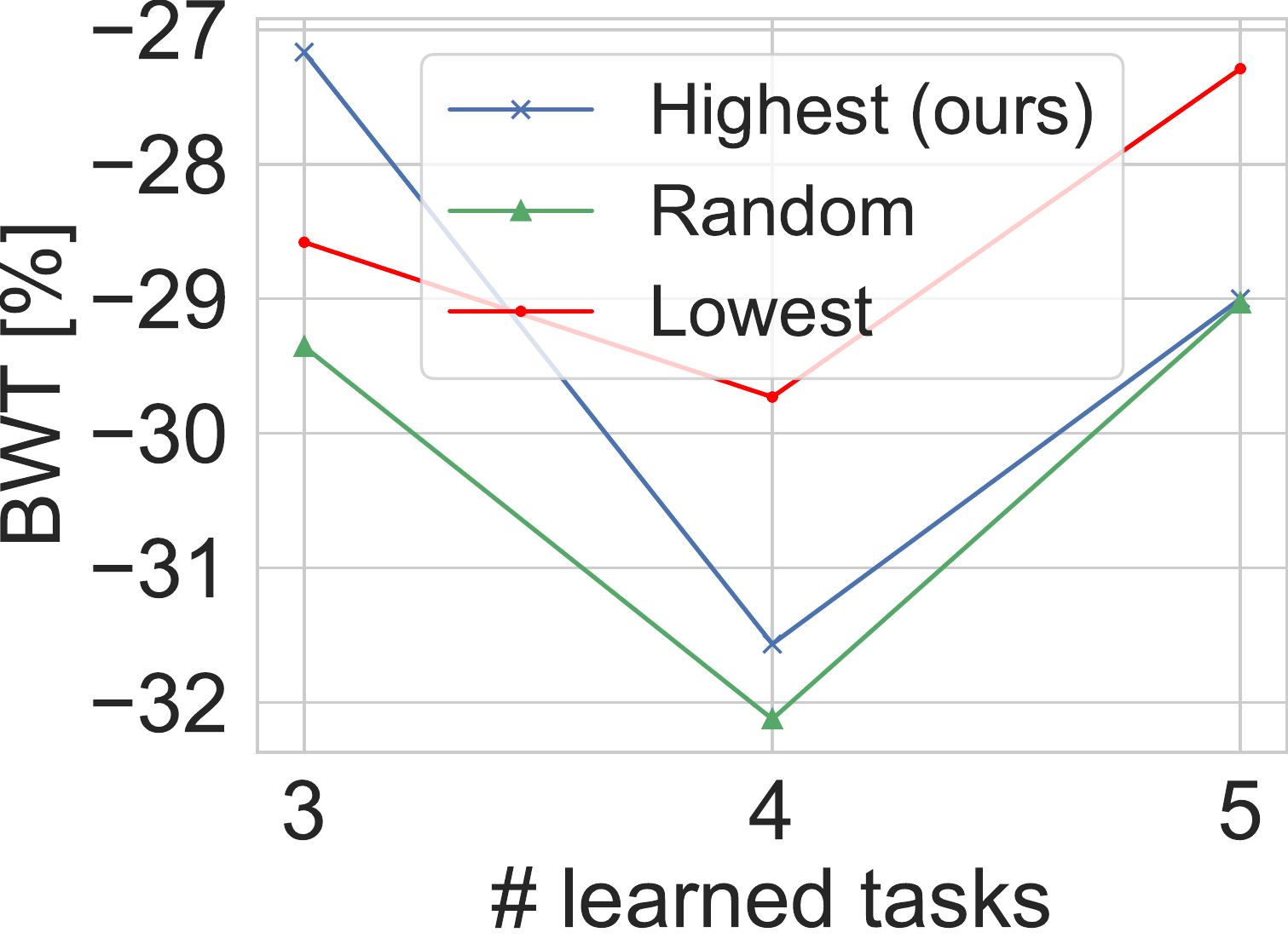} \hfill
    \includegraphics[width=0.3\linewidth]{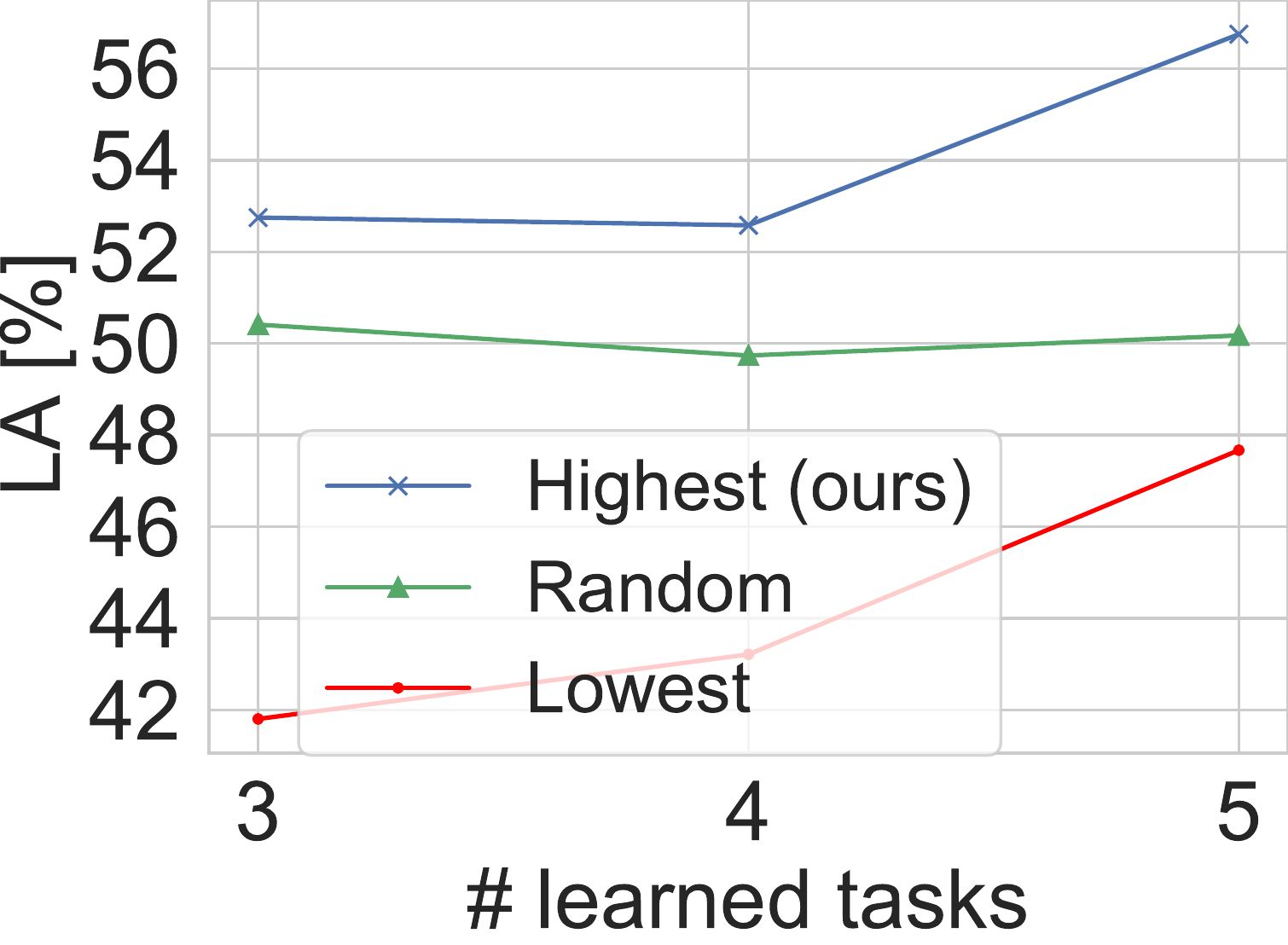}
    \caption{sim-CIFAR10}
    \label{fig:sim-CIFAR_random_lowest}
  \end{subfigure}
  \caption{Performance of AFAF backbone with different strategies for selecting the candidate neurons used to add new connections. }
  \label{fig:analysis_random_lowest}
\end{figure}
\subsection{Analysis}
\label{sec:analysis}
\subsubsection{Effect of Selective Transfer.} To measure the impact of selective transfer and the initially allocated sub-network on forward and backward transfer in AFAF, we compare our selection strategy for the candidate neurons to two other potential strategies discussed in Section \ref{sec:adding_new_componenets}: \textit{Random} and \textit{Lowest}. To reveal the role of selective transfer on performance, we also report the Learning Accuracy (LA) \cite{riemer2018learning}, which is the average accuracy for each task directly after it is learned (Appendix \ref{appendix:experimental_setup}). We calculate this metric starting from the first task that reuses past components in its learning onwards. Figure \ref{fig:analysis_random_lowest} shows the results on Mix datasets and sim-CIFAR10. We present the performance of the tasks that reuse past components (i.e., $t\geq5$ and $t\geq3$ for Mix and sim-CIFAR10, respectively) since the performance of other earlier tasks is the same for all baselines (i.e., the allocation is based on the free neurons). 
As shown in the figure, using the relevant neurons with the highest activation to allocate a new sub-network leads to higher LA on new tasks and lower negative BWT on past tasks than using random neurons. AFAF also has higher ACC than the Random baseline by 4.79\% and 3.01\% on Mix and sim-CIFAR10, respectively. On the other hand, the Lowest baseline limits learning new tasks. It has much lower ACC and LA than the other two baselines. Note that the high BWT of this Lowest baseline is a factor of its low LA. This analysis shows the effect of the \textit{initial} sub-network on performance, although topological optimization occurs during training.       

\begin{table}[t]
\caption{Effect of each contribution in addressing class ambiguities: orthogonal output weights, using free neurons only for allocating output weights, and constraining reusing all past components. ACC is reported in class-IL and task-IL.}
\label{table:class_ambiguity}
\centering
\resizebox{0.7\columnwidth}{!}{
\begin{tabular}{|lc|ccc|c|ccc|}
\hline
 & \hspace{0.2cm}  & \multicolumn{3}{c|}{sim-CIFAR10}& & \multicolumn{3}{c|}{Mix datasets} \\
 \hline
Method&  & class-IL& \hspace{0.2cm} & task-IL& & class-IL& \hspace{0.2cm} & task-IL \\
\hline
AFAF (ours)& & \textbf{45.23$\pm$2.14}& & \textbf{94.57$\pm$0.05}& & \textbf{59.02$\pm$1.76} & &\textbf{93.41$\pm$0.26}  \\\hline
$w/o$ $orth$ $\textbf{W}_L$& & 41.74$\pm$2.05& & 93.20$\pm$0.26& & 56.38$\pm$3.34& & 93.39$\pm$0.30\\
\hline
$w/$ $\mathcal{R}_{L-1}$&  & 40.30$\pm$2.14& &93.27$\pm$0.17& & 56.83$\pm$1.19 & & 92.85$\pm$0.28\\\hline
w/ $l_{reuse} = L-1$&   & 22.27$\pm$1.61& &83.54$\pm$1.64& & 40.98$\pm$2.75 & & 85.05$\pm$1.01\\
\hline
\end{tabular}
}
\end{table}

\subsubsection{Class Ambiguities.} 
\label{sec:class_ambiguities}
We performed an ablation study to assess each of our proposed contributions in addressing class ambiguities in class-IL (Section \ref{sec:class_ambig}). We performed this analysis on Mix datasets and sim-CIFAR10 benchmarks. To show that class ambiguity causes more challenges in class-IL, we also report the performance in task-IL. To compare only the effect of class ambiguities in both scenarios, we assume components-agnostic inference for task-IL. Yet, task labels are used to select the output neurons that belong to the task at hand. 

\textbf{Analysis 1: Effect of orthogonal output weights.} We evaluate a baseline that denotes AFAF without using orthogonal weights in the output layer \enquote{w/o orth $W_L$}. We obtain this baseline by fixing part of the neurons in the last layer instead of fixing all neurons. We use the same fraction used for other fully connected layers (Appendix \ref{appendix:experimental_setup}). Table \ref{table:class_ambiguity} shows the results. Having a large decision margin via orthogonal output weights increases the performance. We observe that the difference between AFAF and this baseline is larger in class-IL than task-IL, which indicates that task-IL is less affected by class ambiguities.  

\textbf{Analysis 2: Effect of using free neurons only in the last layer.} To analyze this effect, we add another baseline that uses free $\mathcal{S}^{free}_{L-1}$ and candidate $\mathcal{R}_{L-1}$ neurons in allocating the output weights. We denote this baseline as \enquote{w/$\mathcal{R}_{L-1}$}. As shown in Table \ref{table:class_ambiguity}, using free neurons only allows learning specific representation that decreases the ambiguities across tasks. The performance in class-IL is improved by 4.93\% and 2.19\% on sim-CIFAR10 and Mix datasets, respectively.  

\textbf{Analysis 3: Effect of constraining reusing all past components.} We analyze the performance obtained by reusing all past components in learning similar tasks (i.e., $l_{reuse}=L-1$). We denoted this baseline as \enquote{w/$l_{reuse}=L-1$}. As shown in Table \ref{table:class_ambiguity}, the performance has decreased dramatically. Despite that the degradation also occurs in task-IL, it has less effect. This shows the challenge of balancing performance, memory, and computational costs in class-IL. 

\begin{figure}
  \centering
  \begin{subfigure}[b]{0.27\columnwidth}
    \includegraphics[width=\linewidth]{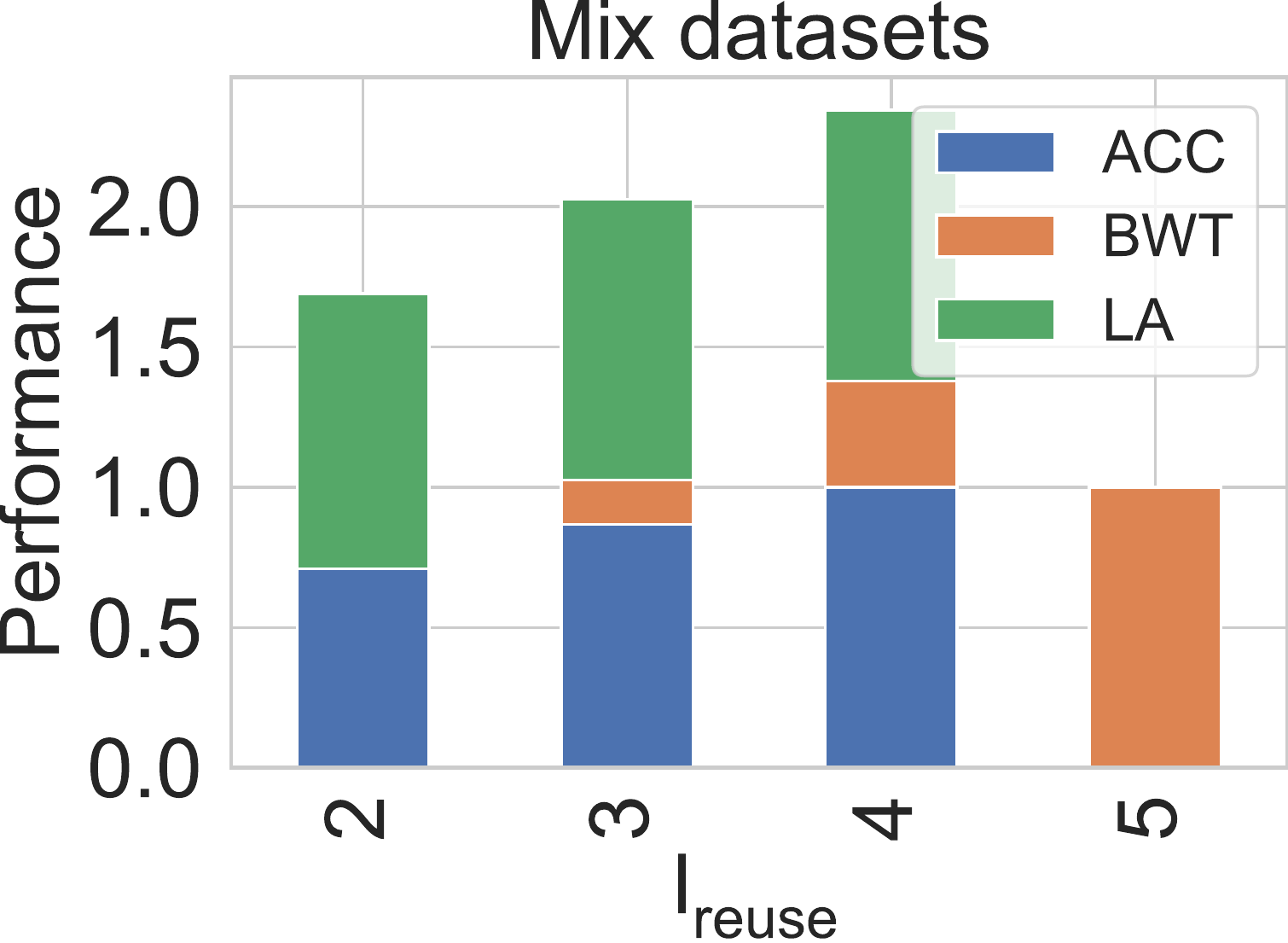} 
    \caption{Mix datasets}
  \end{subfigure}
  \hspace{0.1\columnwidth} 
  \begin{subfigure}[b]{0.27\columnwidth}
    \includegraphics[width=\linewidth]{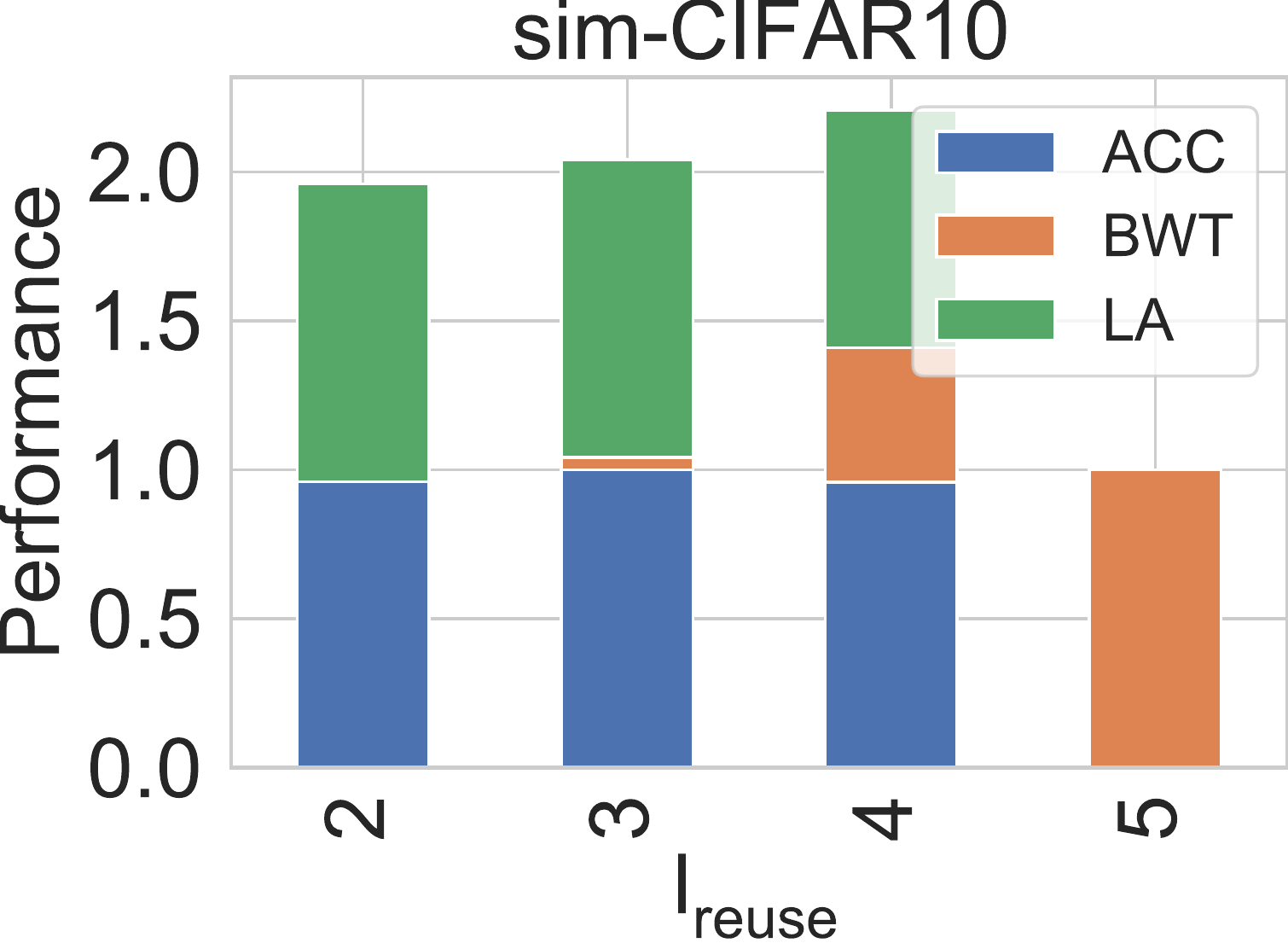} 
    \caption{Sim-CIFAR10}
  \end{subfigure}
  \caption{Normalized performance of AFAF at different values of $l_{reuse}$. }
  \label{fig:analysis_lreuse}
\end{figure}

\subsubsection{Reusable layers.}
We analyze the effect of reusing full layers on performance. We evaluate the performance at the possible values of $l_{reuse} \in [2,L-2]$ (Section \ref{sec:class_ambig}). A min-max scaling is used to normalize the ACC, BWT, and LA; exact values are in Appendix \ref{appendix:reusable_neurons_layers}. Figure \ref{fig:analysis_lreuse} shows the performance of Mix datasets and sim-CIFAR10. Adding new components in lower-level layers ($l_{reuse} \in \{2,3\}$) enables new tasks to achieve high LA but increases forgetting (negative BWT), leading to lower ACC. Reusing the components in lower-level layers while adding new components to learn specific representations in the higher-level ones ($l_{reuse}=4$) achieves a balance between ACC, BWT, and LA. While using a higher value for $l_{reuse}$ limits the performance of new tasks due to class ambiguities, leading to the lowest LA and ACC. 

We performed another study to illustrate the effect of reusing past components in utilizing the fixed-capacity. We show that reusability allows for using higher density for tasks that are dissimilar to the previously learned knowledge, which increases the performance. Details are provided in Appendix \ref{sec:utilize_model}.

\section{Conclusion}
Addressing the stability-plasticity dilemma while balancing CL desiderata is a challenging task. We showed that the challenge increases in the class-IL setting, especially when similar classes are not presented together within the same task. With our proposed task-specific components method, AFAF, we show that altering the model components based on exploiting past knowledge helps in achieving multiple desirable CL properties. Critically, the choice of the sub-network of a new task affects the forward and backward transfer. Hence, we proposed a selection mechanism that selectively transfers relevant knowledge while preserving it. Moreover, we showed that complete layers could be reused in learning similar tasks. Finally, we addressed the class ambiguity that arises in class-IL when similarities increase across tasks and showed that model altering at the connection and neuron levels is more efficient for component-agnostic inference. 
\bibliographystyle{splncs04}
\bibliography{ref}

\begin{thebibliography}{10}
\providecommand{\url}[1]{\texttt{#1}}
\providecommand{\urlprefix}{URL }
\providecommand{\doi}[1]{https://doi.org/#1}

\bibitem{aljundi2018memory}
Aljundi, R., Babiloni, F., Elhoseiny, M., Rohrbach, M., Tuytelaars, T.: Memory
  aware synapses: Learning what (not) to forget. In: Proceedings of the
  European Conference on Computer Vision (ECCV). pp. 139--154 (2018)

\bibitem{atashgahi2021quick}
Atashgahi, Z., Sokar, G., van~der Lee, T., Mocanu, E., Mocanu, D.C., Veldhuis,
  R., Pechenizkiy, M.: Quick and robust feature selection: the strength of
  energy-efficient sparse training for autoencoders. Machine Learning pp. 1--38
  (2021)

\bibitem{bang2021rainbow}
Bang, J., Kim, H., Yoo, Y., Ha, J.W., Choi, J.: Rainbow memory: Continual
  learning with a memory of diverse samples. In: Proceedings of the IEEE/CVF
  Conference on Computer Vision and Pattern Recognition. pp. 8218--8227 (2021)

\bibitem{notMNIST}
Bulatov, Y.: Notmnist dataset. Technical report  (2011),
  \url{http://yaroslavvb.blogspot.it/ 2011/09/notmnist-dataset.html}

\bibitem{chaudhry2018efficient}
Chaudhry, A., Ranzato, M., Rohrbach, M., Elhoseiny, M.: Efficient lifelong
  learning with a-gem. In: International Conference on Learning Representations
  (2018)

\bibitem{chen2020long}
Chen, T., Zhang, Z., Liu, S., Chang, S., Wang, Z.: Long live the lottery: The
  existence of winning tickets in lifelong learning. In: International
  Conference on Learning Representations (2020)

\bibitem{dekhovich2021neural}
Dekhovich, A., Tax, D.M., Sluiter, M.H., Bessa, M.A.: Neural network relief: a
  pruning algorithm based on neural activity. arXiv preprint arXiv:2109.10795
  (2021)

\bibitem{denil2013predicting}
Denil, M., Shakibi, B., Dinh, L., Ranzato, M., de~Freitas, N.: Predicting
  parameters in deep learning. In: Proceedings of the 26th International
  Conference on Neural Information Processing Systems-Volume 2. pp. 2148--2156
  (2013)

\bibitem{dhar2019learning}
Dhar, P., Singh, R.V., Peng, K.C., Wu, Z., Chellappa, R.: Learning without
  memorizing. In: Proceedings of the IEEE/CVF Conference on Computer Vision and
  Pattern Recognition. pp. 5138--5146 (2019)

\bibitem{evci2020rigging}
Evci, U., Gale, T., Menick, J., Castro, P.S., Elsen, E.: Rigging the lottery:
  Making all tickets winners. In: International Conference on Machine Learning.
  pp. 2943--2952. PMLR (2020)

\bibitem{frankle2018lottery}
Frankle, J., Carbin, M.: The lottery ticket hypothesis: Finding sparse,
  trainable neural networks. In: International Conference on Learning
  Representations (2018)

\bibitem{golkar2019continual}
Golkar, S., Kagan, M., Cho, K.: Continual learning via neural pruning. arXiv
  preprint arXiv:1903.04476  (2019)

\bibitem{hadsell2020embracing}
Hadsell, R., Rao, D., Rusu, A.A., Pascanu, R.: Embracing change: Continual
  learning in deep neural networks. Trends in cognitive sciences  (2020)

\bibitem{hoefler2021sparsity}
Hoefler, T., Alistarh, D., Ben-Nun, T., Dryden, N., Peste, A.: Sparsity in deep
  learning: Pruning and growth for efficient inference and training in neural
  networks. Journal of Machine Learning Research  \textbf{22}(241),  1--124
  (2021)

\bibitem{hu2016network}
Hu, H., Peng, R., Tai, Y.W., Tang, C.K.: Network trimming: A data-driven neuron
  pruning approach towards efficient deep architectures. arXiv preprint
  arXiv:1607.03250  (2016)

\bibitem{ioffe2015batch}
Ioffe, S., Szegedy, C.: Batch normalization: Accelerating deep network training
  by reducing internal covariate shift. In: International conference on machine
  learning. pp. 448--456. PMLR (2015)

\bibitem{jayakumar2020top}
Jayakumar, S., Pascanu, R., Rae, J., Osindero, S., Elsen, E.: Top-kast: Top-k
  always sparse training. Advances in Neural Information Processing Systems
  \textbf{33},  20744--20754 (2020)

\bibitem{ke2020continual}
Ke, Z., Liu, B., Huang, X.: Continual learning of a mixed sequence of similar
  and dissimilar tasks. Advances in Neural Information Processing Systems
  \textbf{33} (2020)

\bibitem{kirkpatrick2017overcoming}
Kirkpatrick, J., Pascanu, R., Rabinowitz, N., Veness, J., Desjardins, G., Rusu,
  A.A., Milan, K., Quan, J., Ramalho, T., Grabska-Barwinska, A., et~al.:
  Overcoming catastrophic forgetting in neural networks. Proceedings of the
  national academy of sciences  \textbf{114}(13),  3521--3526 (2017)

\bibitem{krizhevsky2009learning}
Krizhevsky, A., Hinton, G., et~al.: Learning multiple layers of features from
  tiny images. Tech. rep., Citeseer (2009)

\bibitem{krizhevsky2012imagenet}
Krizhevsky, A., Sutskever, I., Hinton, G.E.: Imagenet classification with deep
  convolutional neural networks. Advances in neural information processing
  systems  \textbf{25},  1097--1105 (2012)

\bibitem{lecun1998mnist}
LeCun, Y.: The mnist database of handwritten digits. http://yann. lecun.
  com/exdb/mnist/  (1998)

\bibitem{lee2021sharing}
Lee, S., Behpour, S., Eaton, E.: Sharing less is more: Lifelong learning in
  deep networks with selective layer transfer. In: International Conference on
  Machine Learning. pp. 6065--6075. PMLR (2021)

\bibitem{li2020oslnet}
Li, X., Chang, D., Ma, Z., Tan, Z.H., Xue, J.H., Cao, J., Yu, J., Guo, J.:
  Oslnet: Deep small-sample classification with an orthogonal softmax layer.
  IEEE Transactions on Image Processing  \textbf{29},  6482--6495 (2020)

\bibitem{li2017learning}
Li, Z., Hoiem, D.: Learning without forgetting. IEEE transactions on pattern
  analysis and machine intelligence  \textbf{40}(12),  2935--2947 (2017)

\bibitem{liu2021deep}
Liu, S., Chen, T., Atashgahi, Z., Chen, X., Sokar, G., Mocanu, E., Pechenizkiy,
  M., Wang, Z., Mocanu, D.C.: Deep ensembling with no overhead for either
  training or testing: The all-round blessings of dynamic sparsity. arXiv
  preprint arXiv:2106.14568  (2021)

\bibitem{lopez2017gradient}
Lopez-Paz, D., Ranzato, M.: Gradient episodic memory for continual learning.
  In: Proceedings of the 31st International Conference on Neural Information
  Processing Systems. pp. 6470--6479 (2017)

\bibitem{luo2017thinet}
Luo, J.H., Wu, J., Lin, W.: Thinet: A filter level pruning method for deep
  neural network compression. In: Proceedings of the IEEE international
  conference on computer vision. pp. 5058--5066 (2017)

\bibitem{mallya2018piggyback}
Mallya, A., Davis, D., Lazebnik, S.: Piggyback: Adapting a single network to
  multiple tasks by learning to mask weights. In: Proceedings of the European
  Conference on Computer Vision (ECCV). pp. 67--82 (2018)

\bibitem{mallya2018packnet}
Mallya, A., Lazebnik, S.: Packnet: Adding multiple tasks to a single network by
  iterative pruning. In: Proceedings of the IEEE Conference on Computer Vision
  and Pattern Recognition. pp. 7765--7773 (2018)

\bibitem{masana2020class}
Masana, M., Liu, X., Twardowski, B., Menta, M., Bagdanov, A.D., van~de Weijer,
  J.: Class-incremental learning: survey and performance evaluation. arXiv
  preprint arXiv:2010.15277  (2020)

\bibitem{mazumder2021few}
Mazumder, P., Singh, P., Rai, P.: Few-shot lifelong learning. In: Proceedings
  of the AAAI Conference on Artificial Intelligence. vol.~35, pp. 2337--2345
  (2021)

\bibitem{mccloskey1989catastrophic}
McCloskey, M., Cohen, N.J.: Catastrophic interference in connectionist
  networks: The sequential learning problem. Psychology of learning and
  motivation  \textbf{24},  109--165 (1989)

\bibitem{mermillod2013stability}
Mermillod, M., Bugaiska, A., Bonin, P.: The stability-plasticity dilemma:
  Investigating the continuum from catastrophic forgetting to age-limited
  learning effects. Frontiers in psychology  \textbf{4}, ~504 (2013)

\bibitem{mirzadeh2022architecture}
Mirzadeh, S.I., Chaudhry, A., Yin, D., Nguyen, T., Pascanu, R., Gorur, D.,
  Farajtabar, M.: Architecture matters in continual learning. arXiv preprint
  arXiv:2202.00275  (2022)

\bibitem{mocanu2018scalable}
Mocanu, D.C., Mocanu, E., Stone, P., Nguyen, P.H., Gibescu, M., Liotta, A.:
  Scalable training of artificial neural networks with adaptive sparse
  connectivity inspired by network science. Nature communications
  \textbf{9}(1),  1--12 (2018)

\bibitem{mocanu2016online}
Mocanu, D.C., Vega, M.T., Eaton, E., Stone, P., Liotta, A.: Online contrastive
  divergence with generative replay: Experience replay without storing data.
  arXiv preprint arXiv:1610.05555  (2016)

\bibitem{ozdenizci2021training}
{\"O}zdenizci, O., Legenstein, R.: Training adversarially robust sparse
  networks via bayesian connectivity sampling. In: International Conference on
  Machine Learning. pp. 8314--8324. PMLR (2021)

\bibitem{NEURIPS2020_b4418237}
Raihan, M.A., Aamodt, T.: Sparse weight activation training. In: Larochelle,
  H., Ranzato, M., Hadsell, R., Balcan, M.F., Lin, H. (eds.) Advances in Neural
  Information Processing Systems. vol.~33, pp. 15625--15638. Curran Associates,
  Inc. (2020)

\bibitem{ramasesh2020anatomy}
Ramasesh, V.V., Dyer, E., Raghu, M.: Anatomy of catastrophic forgetting: Hidden
  representations and task semantics. In: International Conference on Learning
  Representations (2020)

\bibitem{rebuffi2017icarl}
Rebuffi, S.A., Kolesnikov, A., Sperl, G., Lampert, C.H.: icarl: Incremental
  classifier and representation learning. In: Proceedings of the IEEE
  conference on Computer Vision and Pattern Recognition. pp. 2001--2010 (2017)

\bibitem{riemer2018learning}
Riemer, M., Cases, I., Ajemian, R., Liu, M., Rish, I., Tu, Y., Tesauro, G.:
  Learning to learn without forgetting by maximizing transfer and minimizing
  interference. In: International Conference on Learning Representations (2018)

\bibitem{rusu2016progressive}
Rusu, A.A., Rabinowitz, N.C., Desjardins, G., Soyer, H., Kirkpatrick, J.,
  Kavukcuoglu, K., Pascanu, R., Hadsell, R.: Progressive neural networks. arXiv
  preprint arXiv:1606.04671  (2016)

\bibitem{serra2018overcoming}
Serra, J., Suris, D., Miron, M., Karatzoglou, A.: Overcoming catastrophic
  forgetting with hard attention to the task. In: International Conference on
  Machine Learning. pp. 4548--4557. PMLR (2018)

\bibitem{shin2017continual}
Shin, H., Lee, J.K., Kim, J., Kim, J.: Continual learning with deep generative
  replay. In: Advances in Neural Information Processing Systems. pp. 2990--2999
  (2017)

\bibitem{sokar2021learning}
Sokar, G., Mocanu, D.C., Pechenizkiy, M.: Learning invariant representation for
  continual learning. In: Meta-Learning for Computer Vision Workshop at the
  35th AAAI Conference on Artificial Intelligence (AAAI-21) (2021)

\bibitem{sokar2021self}
Sokar, G., Mocanu, D.C., Pechenizkiy, M.: Self-attention meta-learner for
  continual learning. In: Proceedings of the 20th International Conference on
  Autonomous Agents and MultiAgent Systems. pp. 1658--1660 (2021)

\bibitem{sokar2021spacenet}
Sokar, G., Mocanu, D.C., Pechenizkiy, M.: Spacenet: Make free space for
  continual learning. Neurocomputing  \textbf{439},  1--11 (2021)

\bibitem{sokar2021dynamic}
Sokar, G., Mocanu, E., Mocanu, D.C., Pechenizkiy, M., Stone, P.: Dynamic sparse
  training for deep reinforcement learning. In: International Joint Conference
  on Artificial Intelligence (2022)

\bibitem{van2020brain}
van~de Ven, G.M., Siegelmann, H.T., Tolias, A.S.: Brain-inspired replay for
  continual learning with artificial neural networks. Nature communications
  \textbf{11}(1),  1--14 (2020)

\bibitem{veniat2021efficient}
Veniat, T., Denoyer, L., Ranzato, M.: Efficient continual learning with modular
  networks and task-driven priors. In: International Conference on Learning
  Representations (2021)

\bibitem{wortsman2020supermasks}
Wortsman, M., Ramanujan, V., Liu, R., Kembhavi, A., Rastegari, M., Yosinski,
  J., Farhadi, A.: Supermasks in superposition. Advances in Neural Information
  Processing Systems  \textbf{33} (2020)

\bibitem{xiao2017fashion}
Xiao, H., Rasul, K., Vollgraf, R.: Fashion-mnist: a novel image dataset for
  benchmarking machine learning algorithms. arXiv preprint arXiv:1708.07747
  (2017)

\bibitem{yoon2019scalable}
Yoon, J., Kim, S., Yang, E., Hwang, S.J.: Scalable and order-robust continual
  learning with additive parameter decomposition. In: International Conference
  on Learning Representations (2019)

\bibitem{yoon2018lifelong}
Yoon, J., Yang, E., Lee, J., Hwang, S.J.: Lifelong learning with dynamically
  expandable networks. In: International Conference on Learning Representations
  (2018)

\bibitem{zenke2017continual}
Zenke, F., Poole, B., Ganguli, S.: Continual learning through synaptic
  intelligence. In: International Conference on Machine Learning. pp.
  3987--3995. PMLR (2017)

\bibitem{zhu2019multi}
Zhu, H., Jin, Y.: Multi-objective evolutionary federated learning. IEEE
  Transactions on Neural Networks and Learning Systems  \textbf{31}(4),
  1310--1322 (2019)

\end{thebibliography}
\clearpage
\appendix

\section{Experimental Settings}
\label{appendix:experimental_setup}
\subsection{Benchmarks}
CIFAR-10 \cite{krizhevsky2009learning} is a well-known dataset for classification tasks. It contains tiny natural images of size ($32\times32$). It consists of 10 classes and has 6000 images per class (6000 training + 1000 test). We use this dataset to construct two benchmarks: split-CIFAR10 and sim-CIFAR10. Split-CIFAR10 is the standard benchmark for CL. It consists of 5 tasks, and each task has two classes. The benchmark has the typical class order of CIFAR10 (Table \ref{table:details_split_CIFAR10}). To assess the performance of a model under more challenging conditions, we construct sim-CIFAR10 with a higher similarity between classes across tasks. This is obtained by shuffling the class order, as shown in Table \ref{table:details_sim_CIFAR10}.

CIFAR-100 \cite{krizhevsky2009learning} is a more complex dataset with 100 classes and fewer samples for each. It has 600 images per class (500 train + 100 test). The images have the same size as those in CIFAR10. Classes within the same superclass in this dataset have semantic similarities. We exploit this property to construct a new benchmark with a high similarity level across tasks. We called this benchmark sim-CIFAR100. We choose eight superclasses and two classes from each superclass. We distribute  the classes from the same superclass in different tasks, constructing 8 tasks. The class order and the superclasses used in each task are illustrated in Table \ref{table:details_sim-CIFAR100}.   

MNIST dataset \cite{lecun1998mnist} contains grayscale images of size $28 \times 28$, representing hand-written digits from 0 to 9. It has 60,000 training images and 10,000 test images.

Fashion-MNIST dataset \cite{xiao2017fashion} is more complex than MNIST. The images show individual articles of clothing. It has the same sample size and structure of training and test sets as MNIST.

NotMNIST dataset \cite{notMNIST} is similar to MNIST. It has 10 classes, with letters A-J taken from different fonts.

Following \cite{serra2018overcoming,veniat2021efficient}, we used CIFAR10, MNIST, Fashion MNIST, and NotMNIST datasets to construct a sequence of tasks from different domains, named Mix datasets. We split each dataset into two parts of five classes each. We organize the tasks such that the model sees all dissimilar tasks before it encounters new tasks similar to the previously seen ones. Table \ref{table:details_mix} shows the details of the task order and the classes in each task.

\subsection{Network architecture} We follow \cite{serra2018overcoming,ke2020continual,veniat2021efficient} to use an AlexNet-like architecture \cite{krizhevsky2012imagenet}. The details of the network architecture are in Table \ref{network_architecture}. For sim-CIFAR100, we replaced the dropout layers with batch normalization \cite{ioffe2015batch} since it was more effective in reducing the overfitting to the small number of samples per class in this dataset. Moreover, for regularization-based methods, batch normalization achieved higher accuracy than dropout layers. Hence, we used batch normalization for these methods in all benchmarks except for Mix datasets since the tasks come from different domains. 

\begin{table}[H]
\caption{Details of the tasks in the split-CIFAR10 benchmark. }
\label{table:details_split_CIFAR10}
\begin{center}
\begin{tabular}{|c|c|c|}
\hline
{Task id} & {Classes} & {Category}
\\ \hline
1 & [airplane, car]  &  vehicle  \\
2 & [bird, cat] & animal\\ 
3 & [deer, dog]  & animal \\
4 & [frog, horse] & animal\\
5 & [ship, truck] & vehicle\\
\hline
\end{tabular}
\end{center}
\end{table}

\begin{table}[H]
\caption{Details of the tasks in the sim-CIFAR10 benchmark. }
\label{table:details_sim_CIFAR10}
\begin{center}
\begin{tabular}{|c|c|c|}
\hline
{Task id} & {Classes} & {Category}
\\ \hline
1 & [car, cat]  & animal, vehicle  \\
2 & [horse, truck] & animal, vehicle \\ 
3 & [dog, deer]  & animal \\
4 & [airplane, bird] & animal, vehicle\\
5 & [frog, ship] & animal, vehicle\\
\hline
\end{tabular}
\end{center}
\end{table}

\begin{table}[H]
\caption{Details of the tasks in the sim-CIFAR100 benchmark. }
\label{table:details_sim-CIFAR100}
\begin{center}
\resizebox{0.7\columnwidth}{!}{
\begin{tabular}{|c|c|c|}
\hline
{Task id} & {Classes} & {Category (Superclass)}
\\ \hline
1 & [apples, girl] & fruit, people \\
2 & [mouse, bicycle] & small mammals, vehicles \\
3 & [bee, lion] & insects, large carnivores\\
4 & [bottle, couch] & food containers, household furniture	\\
5 & [orange, boy] & fruit, people \\ 
6 & [rabbit, motocycle] & small mammals, vehicles\\
7 & [butterfly, tiger] & insects, large carnivores \\
8 & [cans, chair] & food containers, household furniture\\
\hline
\end{tabular}
}
\end{center}
\end{table}

\begin{table}[h]
\caption{Details of the tasks in the Mix datasets benchmark. }
\label{table:details_mix}
\begin{center}
\resizebox{0.7\columnwidth}{!}{
\begin{tabular}{|c|c|c|}
\hline
{Task id}  & {Classes} & {dataset} \\ \hline
1   & [airplane, car, bird, cat, deer] & CIFAR10 \\
2 & [0, 1, 2, 3, 4] & MNIST  \\ 
3 & [A, B, C, D, E] & NotMNIST\\
4  & [T-shirt/top, trouser, pullover, dress, coat]& FashionMNIST\\
5 & [dog, frog, horse, ship, truck] & CIFAR10 \\
6  & [5,6,7,8,9] & MNIST \\
7  & [F, G, H, I , J] & NotMNIST\\
8 & [sandal, shirt, sneaker, bag, ankle boot] & FashionMNIST \\
\hline
\end{tabular}
}
\end{center}
\end{table}

\begin{table}[t]
\caption{Network Architecture.}
\label{network_architecture}
\begin{center}
\resizebox{0.4\columnwidth}{!}{
\begin{tabular}{|l|c|c|c|r|}
\hline
Layer & Channel & Kernel & Dropout \\
\hline
32 $\times$ 32 input & 3 & 3 $\times$ 3 &  \\ \hline
Conv 1 & 64 & 3 $\times$ 3 & \\
Conv 2 & 128 & 3 $\times$ 3 & \\
MaxPool &    &  2 $\times$ 2 & 0.2 \\
Conv 3 & 256 & 3 $\times$ 3 & \\
MaxPool &    &  2 $\times$ 2 &  \\
Dense 1 & 2048 &   & 0.2 \\
Dense 2 & 2048 &   &  \\ \hline
Softmax Output & $C \times T$& & \\
\hline
\end{tabular}
}
\end{center}

\end{table}

\begin{table*}[t]
\caption{Percentage of allocated $h_l^{alloc}$ and fixed neurons in convolution layers (conv), fully connected layers (fc), and output layer. The density of the added connections with respect to the allocated neurons is also reported.}
\label{table:alloc_precentage}
\centering
\begin{tabular}{|l|ccc|ccc|ccc|}
\hline
 & \multicolumn{3}{c|} {$h_l^{alloc}$[\%]}  & \multicolumn{3}{c|} {fixed[\%]} & \multicolumn{3}{c|} {density level [\%]} \\
 \hline
Benchmark & conv & fc & output & conv  & fc &  output &  conv  &  fc &  output \\
\hline
Split-CIFAR10 & 70 & 20 & 10 & 10 & 30 & 100 & 25& 25& 70\\
Sim-CIFAR10 & 50 & 50 & 20 & 20 & 20 & 100 & 30 & 20 & 70 \\
Sim-CIFAR100 & 30 & 20 & 10 & 10 & 30 & 100 & 20 & 40 & 80\\
Mix datasets & 70 & 10 & 10 & 10 & 50 & 100 & 10 & 20 & 90 \\
\hline
\end{tabular}
\end{table*}

\subsection{Implementation Details} 
All input images are resized to $32\times32$. Gray-scale images are converted to 3 channels. The network is trained using stochastic gradient descent with a batch size of 64 and a learning rate of 0.1. Each task is trained for 40 epochs. The hyperparameters are selected using a random search. While our proposed method, AFAF, is capable of coping with both structured and unstructured sparsity \cite{hoefler2021sparsity}, we decided to implement it using structured sparsity to enable a quick adoption in the near future of its computational benefits.

We use $l_{reuse}$ of 4 for all benchmarks (i.e., the convolution layers are reused in learning future tasks). We start reusing previous components in learning future tasks after acquiring similar tasks. Therefore, for Mix datasets and sim-CIFAR100 benchmarks, we start reusing from task 5, while for split-CIFAR10 and sim-CIFAR10, we start from task 3.
To allocate a new topology, a set of neurons is selected in each layer $h_l^{alloc}$. For all benchmarks, 70\% of $h_l^{alloc}$ are free neurons $S_l^{free}$ and 30\% are candidates $\mathcal{R}^c_l$ (Analysis is in Appendix \ref{appendix:reusable_neurons_layers}). Table \ref{table:alloc_precentage} shows the percentage of allocated neurons in each layer, the percentage of fixed neurons after training, and the density level of the connections allocated for all benchmarks. Note the density level is reported with respect to the allocated number of neurons.

For SpaceNet, the percentage of $h_l^{alloc}$, fixed neurons, and density levels are the same as the ones used for AFAF. Connections are allocated in each layer for each task using the free neurons. 

For regularization-based methods, EWC~\cite{serra2018overcoming}, MAS~\cite{aljundi2018memory}, and LWF~\cite{li2017learning}, we used the public code from~\cite{masana2020class} to produce results on the studied benchmarks. We use a regularization factor of 5000 and 1 for EWC and MAS, respectively.

For PackNet, we use the official code from \cite{mallya2018packnet}. Note that PackNet was originally designed for task-IL, as discussed in Section \ref{sec:related_work}. To adapt PackNet to class-IL, we use all the learned connections during inference without masks. A dense model is trained from scratch on the CL tasks. After learning each task, a percentage of unimportant weights is pruned. We select this percentage such that all tasks have the same sparsity level. Each task is trained for 40 epochs, and another 20 epochs are performed after pruning to restore the performance. The sparsity level used for each task is 10\% for Mix datasets and sim-CIFAR100 benchmarks. For split-CIFAR10 and sim-CIFAR10, a sparsity level of 20\% is used. We tried using a higher sparsity level for PackNet. However, this results in lower performance.

\subsection{Evaluation Metrics}
\textbf{Average Accuracy (ACC).} The average accuracy after a model has been trained sequentially till task $T$. 
\begin{equation}
    ACC = \sum_{i=1}^{T} a_{T,i},
\end{equation}
where $a_{j,i}$ is the accuracy on task $i$ after learning the $j$-th task in the sequence, and $T$ is the total number of seen tasks.

\textbf{Backward transfer (BWT)} \cite{lopez2017gradient}. This metric measures the influence of learning new tasks on the performance of previous tasks. Formally BWT is calculated as follows:
\begin{equation}
\label{BWT}
\begin{split}
   BWT = \frac{1}{T-1} \sum_{i=1}^{T-1} a_{T,i} - a_{i,i}. 
\end{split}
\end{equation}
Larger negative backward transfer indicates catastrophic forgetting.

\textbf{Learning Accuracy (LA)} \cite{riemer2018learning}. This metric measures the average accuracy for each task directly after it is learned as follows: 
\begin{equation}
\label{LA}
\begin{split}
   LA = \frac{1}{T} \sum_{i=1}^T a_{i,i}.  
\end{split}
\end{equation}
\textbf{Model size (\# param)}. This metric estimates the memory cost consumed by a CL model. The network size is estimated by the summation of the number of connections allocated in its layers as follows:
\begin{equation}
    \#params = \sum_{l=1}^{L} \norm{\bm{W}_{l}}_{0},
\end{equation}
where $\bm{W}_{l}$ is the actual weights used in layer $l$ after the model learns all tasks, $\norm{.}_0$ is the standard $L_0$ norm, and $L$ is the number of layers in a model. For sparse networks, $\norm{\bm{W}_{l}}_{0}$ is controlled by the sparsity level of each task.

\textbf{Floating-point operations (FLOPs).} This metric estimates the computational cost of a method by calculating how many FLOPs are required for training. We follow the method described in \cite{evci2020rigging} to calculate the FLOPs. The FLOPs are calculated with the total number of multiplications and additions layer by layer in the network that occurs during forward and backward passes. 

Let $f_D$ be the number of FLOPs required to train a dense model on the sequence of tasks. The FLOPs required to train a model with sparse subnetworks is $f_s \approx (1-s) \times f_D$, where $s$ is the sparsity level of the model after learning all tasks.
The FLOPs of task $t$ using PackNet that trains dense connections and prunes the model after convergence is $f_D\times d_{t}$ + $f_t^{tune}$, where $d_{t}$ is the density of the trained connections at time $t$ and $f_t^{tune}$ is the number of FLOPs required to finetune each task after pruning. Note that some regularization methods, such as EWC \cite{kirkpatrick2017overcoming}, require additional FLOPs to calculate the regularization loss. Here, we omit this cost, focusing on the computational costs resulting from training either dense or sparse networks.  

\section{Details of Illustrative Experiments}
\label{sec:illustrative_experiments}
In this appendix, we give the details of the illustrative experiments in Section \ref{sec:intro}. The experiments are performed on two tasks, A and B, constructed from CIFAR10. Task A has two classes, \{cat, car\}, while Task B has classes of \{dog, truck\}. We use the same experimental settings of sim-CIFAR10 and network architecture described in Appendix \ref{appendix:experimental_setup}. 

\textbf{Analysis 1:} The effect of the allocated sub-network on performance. Figure \ref{fig:initial_topology} shows the performance using AFAF (\enquote{Sub-network 1}), and the lowest baseline described in Section \ref{sec:method} (\enquote{Sub-network 2}) for allocating the sparse topology of Task B. Different initial sub-networks leads to different behavior and could either balance or be biased toward one of CL requirements.  

\textbf{Analysis 2:} Using the same altering of the model components in task-IL and class-IL. Figure \ref{fig:acc_TIL_CIL} shows the performance when all the learned components of Task A are reused in learning Task B (i.e., $l_{reuse} = L-1$). 

\textbf{Analysis 3:} Addition of new components versus reusing existing ones. In this experiment, sparse connections are allocated for Task A. Two baselines are evaluated. One baseline allocates new sparse connections in each layer for Task B (i.e., applying the SpaceNet method). The other baseline represents AFAF. It reuses the previous components and starts allocating new connections from $l_{reuse}$. The results of this analysis are illustrated in Figure \ref{fig:add_reuse}.

\begin{table*}[t]
\caption{ACC of task-specific component methods in task-IL and class-IL on sim-CIFAR10 and Mix datasets. }
\label{table:analysis_connection_vs_neuron}
\begin{center}
\resizebox{\columnwidth}{!}{
\begin{tabular}{|l|c|cc|cc|}
\hline
& & \multicolumn{2}{c}{sim-CIFAR10}& \multicolumn{2}{|c|}{Mix datasets}\\
\hline
{Strategy} & {Method} & {Task-IL} & {Class-IL} & {Task-IL} & {Class-IL} 
\\ \hline
Connection level &  PackNet & 89.33$\pm$0.02 & 32.46$\pm$1.22 & \textbf{96.52$\pm$0.05} & 16.61$\pm$2.35\\
Connection and neuron levels & AFAF & \textbf{94.44$\pm$0.28} & \textbf{45.23$\pm$2.14} &  93.41$\pm$0.26 & \textbf{59.01$\pm$1.76}  \\
\hline
\end{tabular}
}
\end{center}
\end{table*}

\section{Connection Level versus Neuron Level Altering}
\label{appendix:connection_versus_neuron}
Section \ref{sec:results} showed that methods that alter the model components based on the connection and neuron levels are more efficient than the ones that alter at the connection level only (i.e., PackNet). To reveal that altering at the two levels is more necessitous in class-IL, we illustrate the performance of PackNet in task-IL. Table \ref{table:analysis_connection_vs_neuron} shows the results on sim-CIFAR10 and Mix datasets. 

PackNet achieves a good performance in task-IL despite the low performance achieved in class-IL, especially on the Mix datasets benchmark. This illustrates the challenges that arise from operating in the class-IL setting without access to task labels at inference. Reducing the interference between representations is more effective in this setting. 

AFAF achieves comparable performance to PackNet in task-IL with reusing previous components and selective transfer. More interestingly, AFAF does not store a mask for each task to select the specific components at inference (i.e., component-agnostic inference). This leads to more efficient CL models with less memory and computation costs. 

Note that the hyper-parameters (e.g., sparsity level, fixed neuron percentage, etc.) of AFAF are not tuned for task-IL. We used the same hyperparameters as for class-IL. Higher performance might be achieved using hyperparameter optimization for task-IL. Yet, it is not the focus of this paper.

\begin{figure}
  \centering
  \begin{subfigure}[b]{0.3\columnwidth}
    \includegraphics[width=\linewidth]{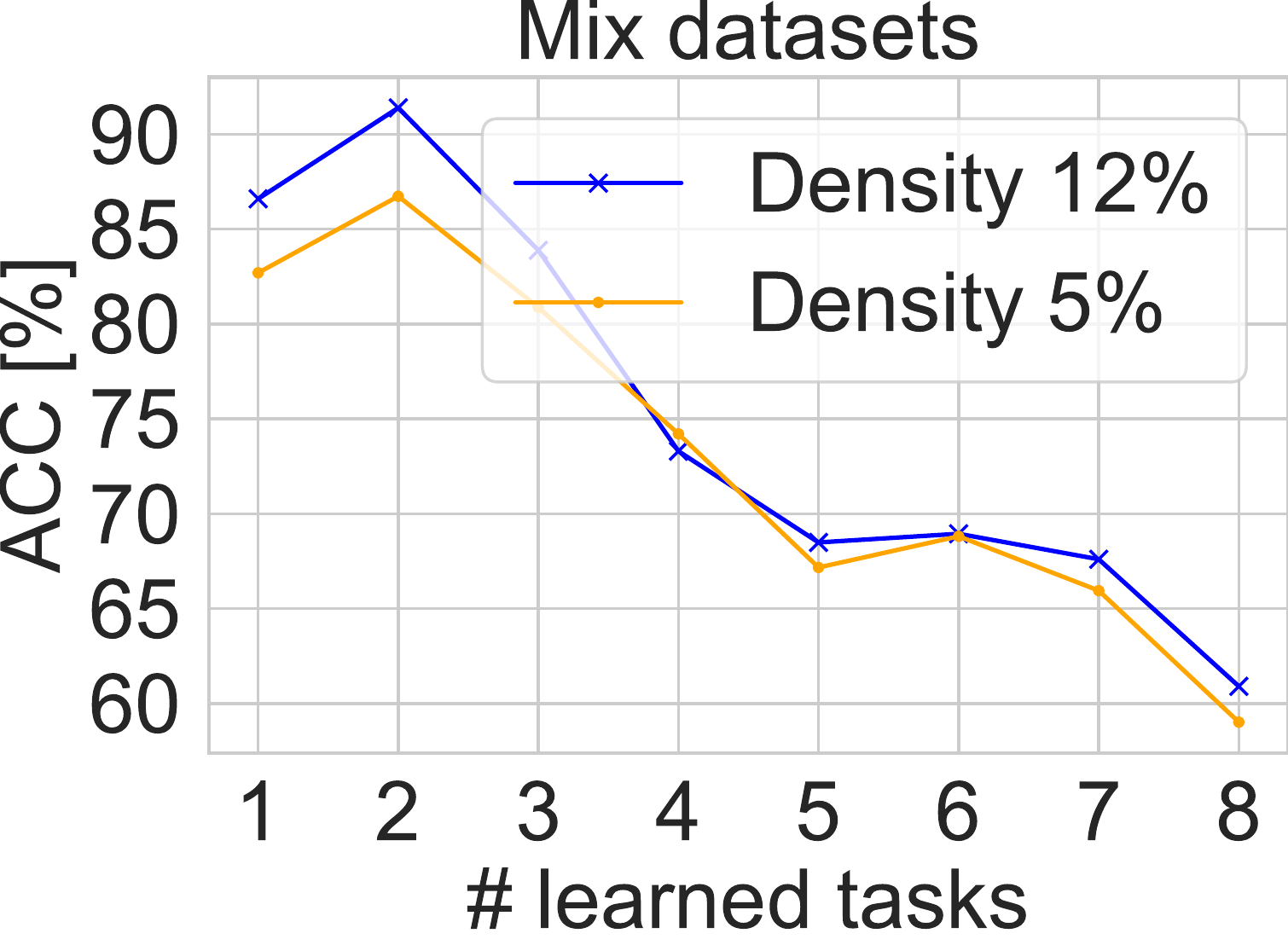} 
    \caption{ACC ($\uparrow$)}
    \label{fig:ACC_high_density}
  \end{subfigure}
  \hspace{0.1\columnwidth} 
  \begin{subfigure}[b]{0.3\columnwidth}
    \includegraphics[width=\linewidth]{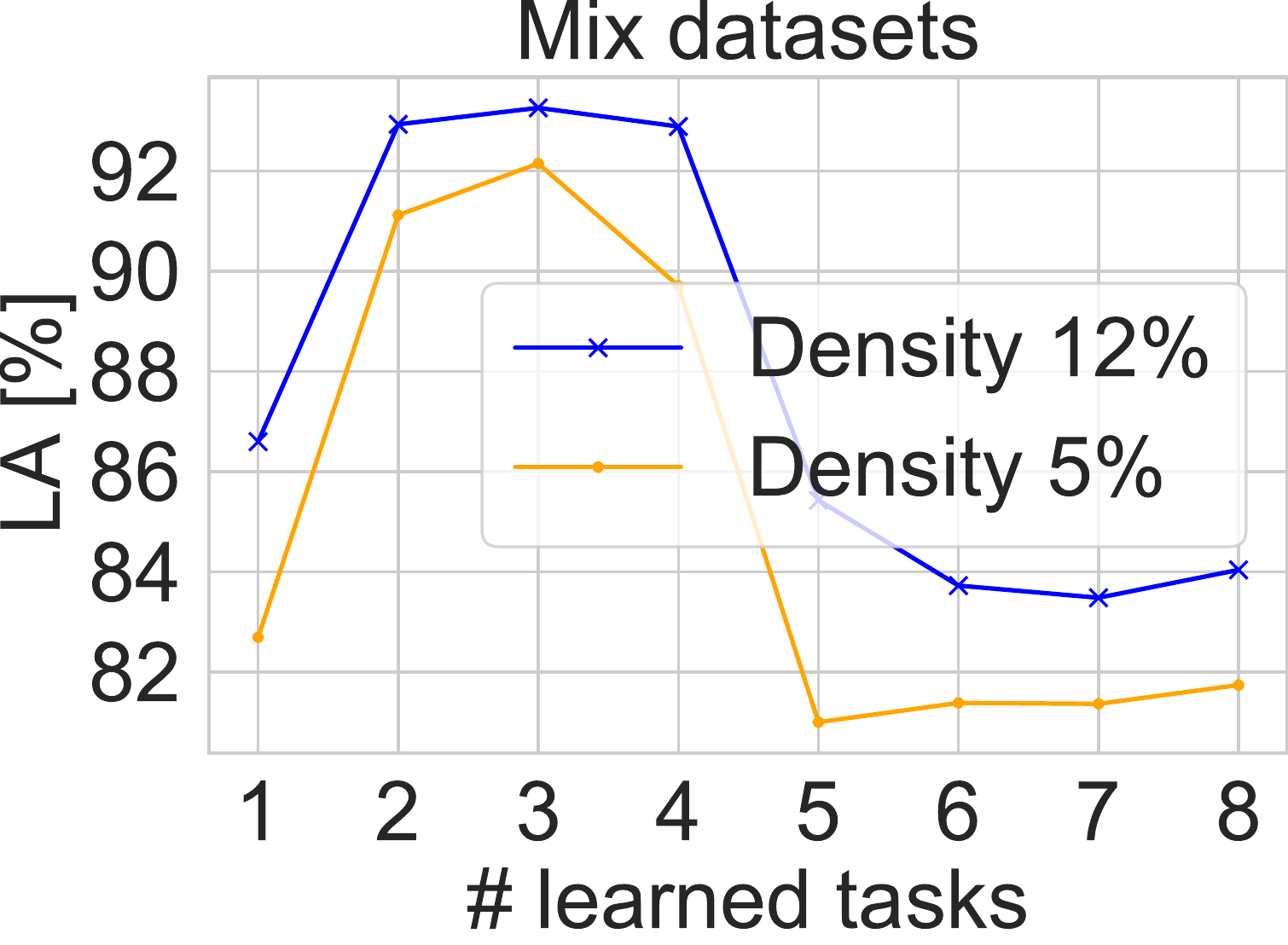} 
    \caption{LA ($\uparrow$)}
    \label{fig:LA_high_density}
  \end{subfigure}
  \caption{Performance of AFAF at different density levels of sparse sub-networks of the first four tasks of Mix datasets benchmark. }
  \label{fig:analysis_high_density}
\end{figure}

\section{Utilizing Fixed-capacity Models}
\label{sec:utilize_model}\
Using fixed-capacity models constrains the sparsity level of the sub-network used for each task. Very high sparse models might limit the performance of complex tasks. Here, we illustrate that reusing previous components in learning similar tasks allows dissimilar ones to have a higher density. We performed this analysis on Mix datasets. To this end, for the first four dissimilar tasks in this benchmark, we increased the density level for the layers that will be reused in the second four tasks (i.e., $l < l_{reuse}$). The new density for each task in this baseline is 12\%, while the one used to obtain the main results in Section \ref{sec:results} is 5\% (Appendix \ref{appendix:experimental_setup}). Figure \ref{fig:analysis_high_density} shows the ACC and LA for each case. Using higher density for the dissimilar tasks that are not encountered before increases their performance. The achieved ACC and LA across all time steps is higher than using a density of 5\% for each task. Without reusability, as in SpaceNet, a model could not fit all tasks with this higher density. This reveals the importance of utilizing the available capacity efficiently. 

\section{Analysis of Reusable Layers and Neurons}
\label{appendix:reusable_neurons_layers}
\begin{figure}
  \centering
  \begin{subfigure}[b]{0.23\columnwidth}
    \includegraphics[width=1.\linewidth]{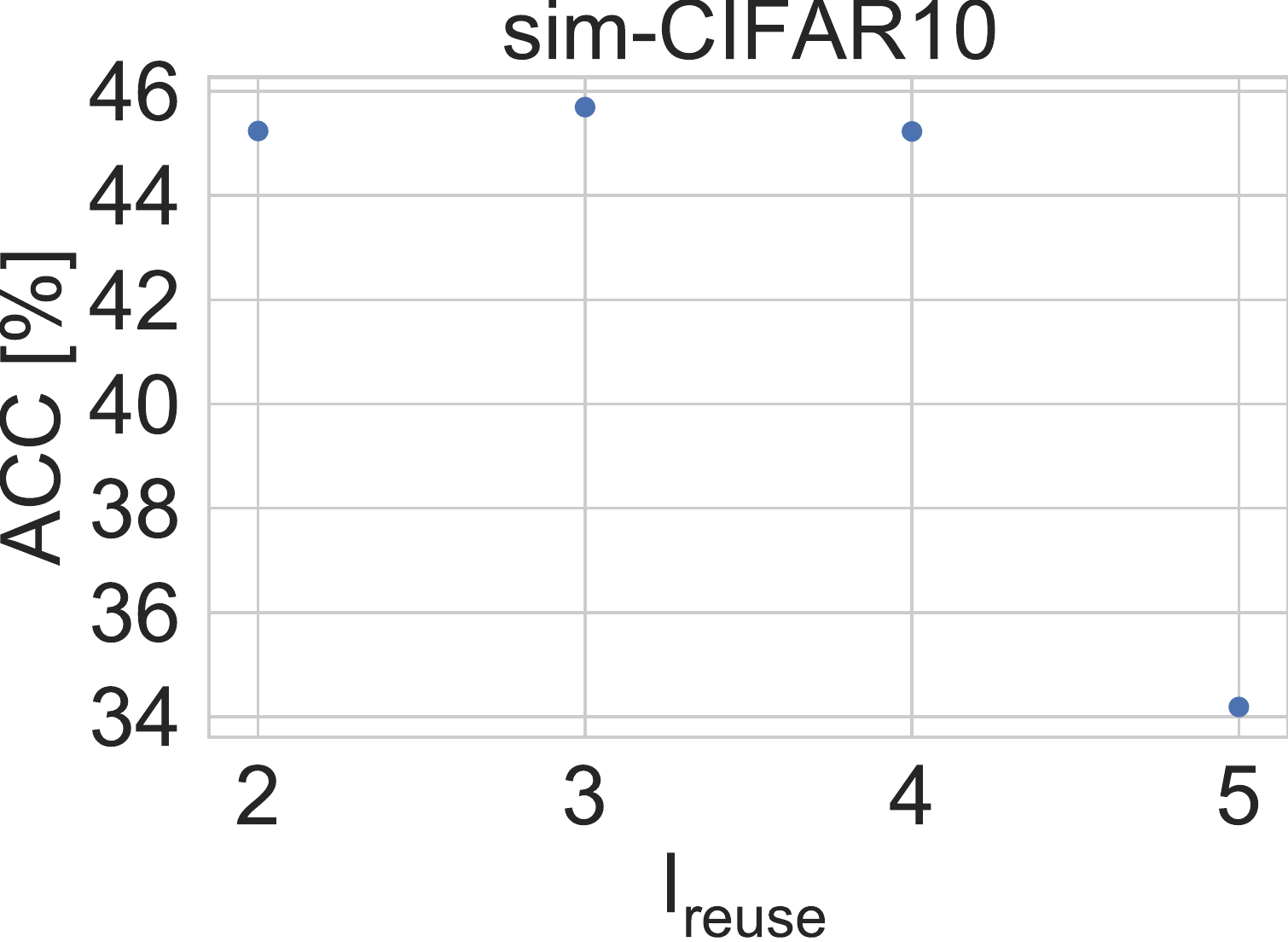} 
    \caption{ACC}
  \end{subfigure}
  \begin{subfigure}[b]{0.23\columnwidth}
    \includegraphics[width=1.\linewidth]{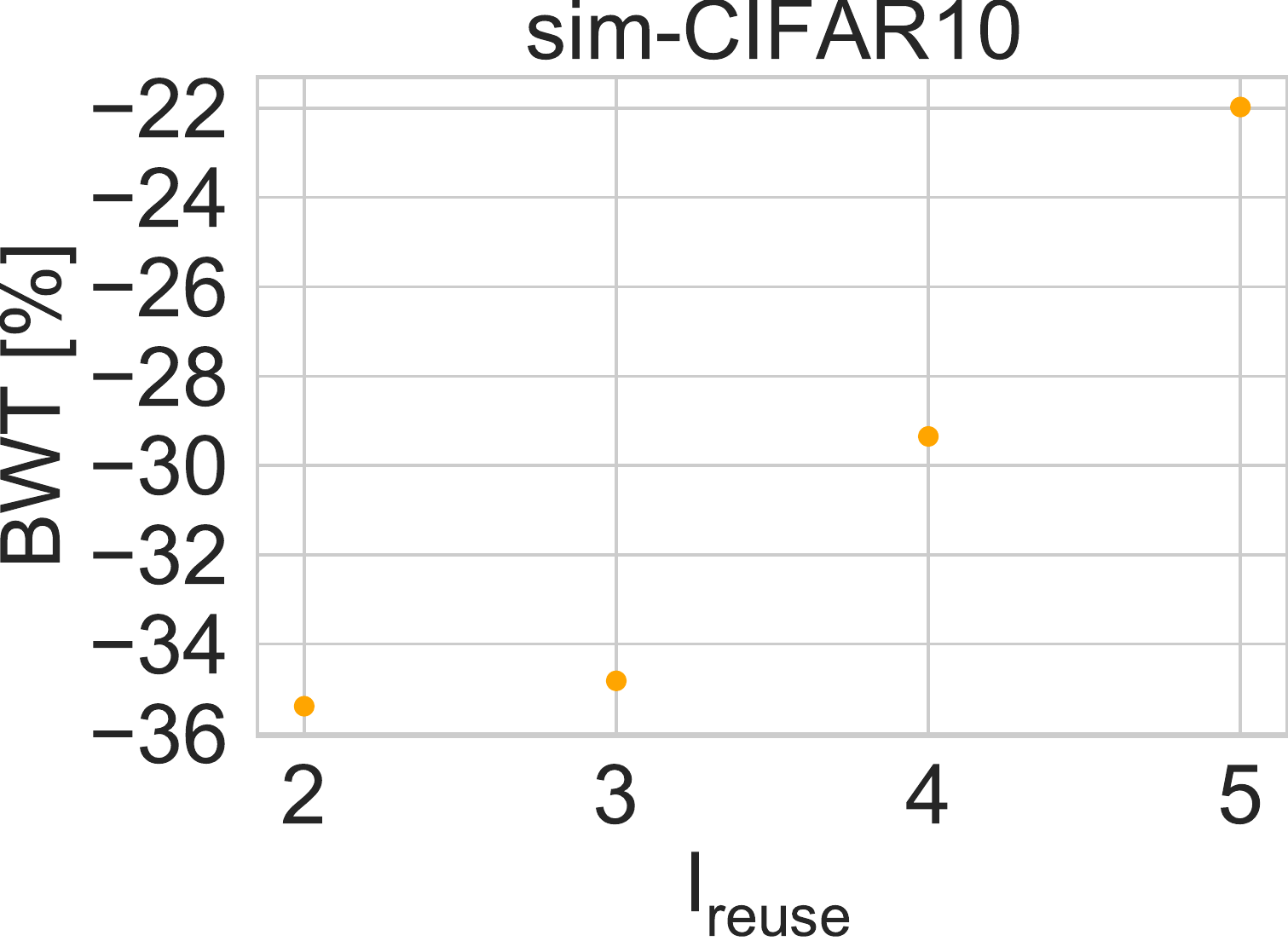} 
    \caption{BWT}
  \end{subfigure}
    \begin{subfigure}[b]{0.23\columnwidth}
    \includegraphics[width=1.\linewidth]{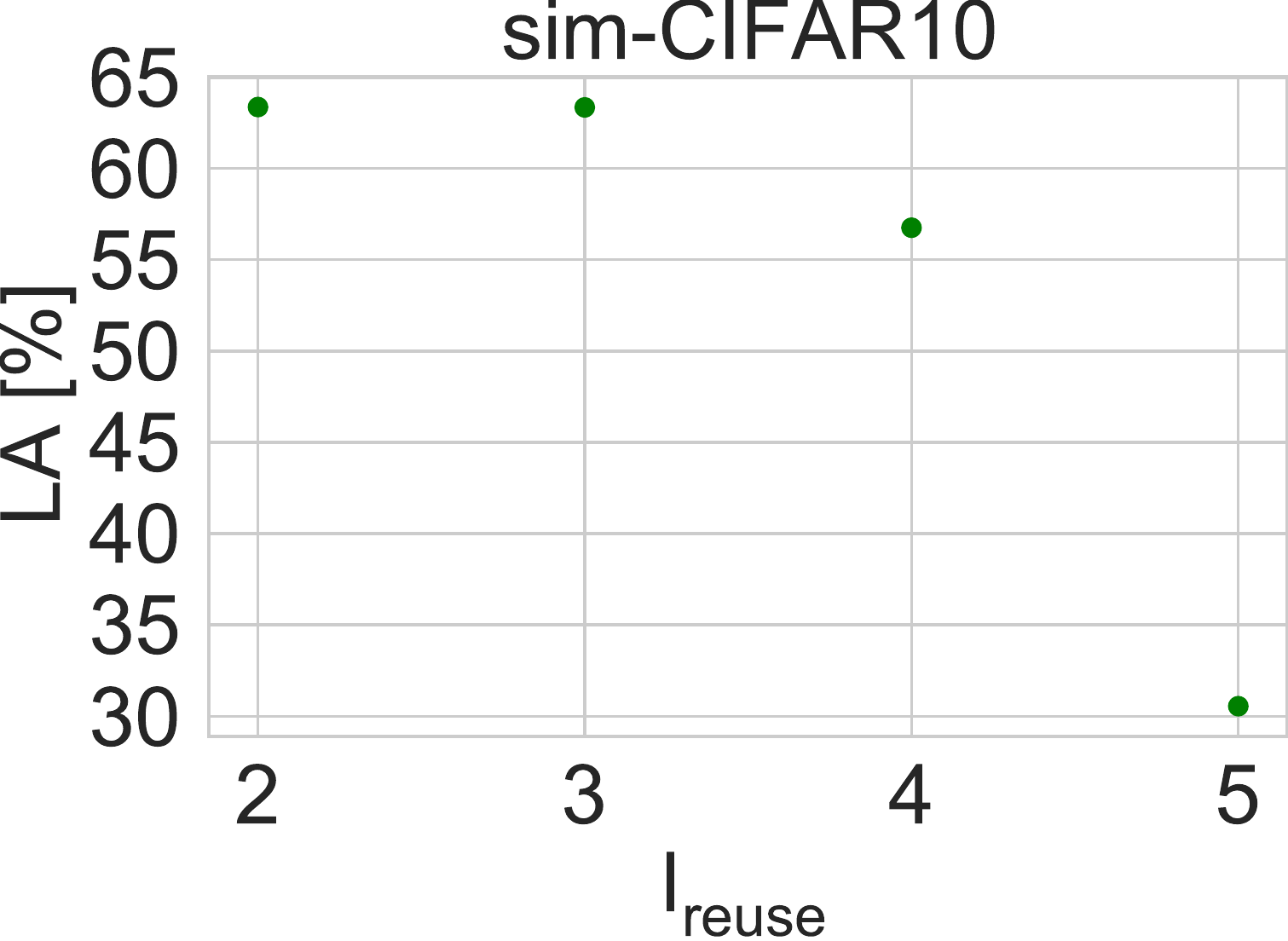} 
    \caption{LA}
  \end{subfigure}
    \begin{subfigure}[b]{0.23\columnwidth}
    \includegraphics[width=1.\linewidth]{pic/sim-cifar10-lreuse_performance_.pdf} 
    \caption{Performance}
  \end{subfigure}
  
    \centering
  \begin{subfigure}[b]{0.23\columnwidth}
    \includegraphics[width=1.\linewidth]{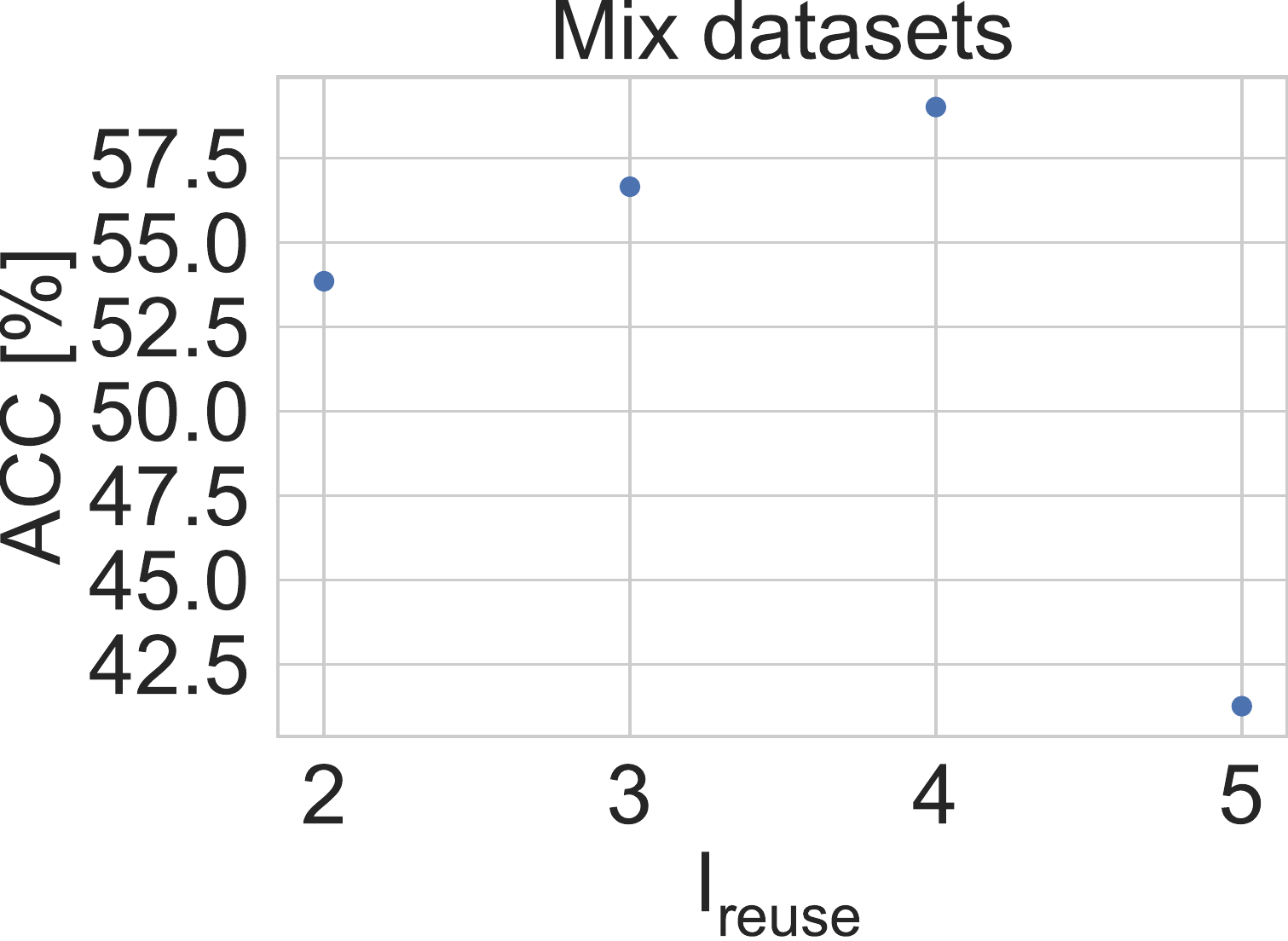} 
    \caption{ACC}
  \end{subfigure}
  \begin{subfigure}[b]{0.23\columnwidth}
    \includegraphics[width=1.\linewidth]{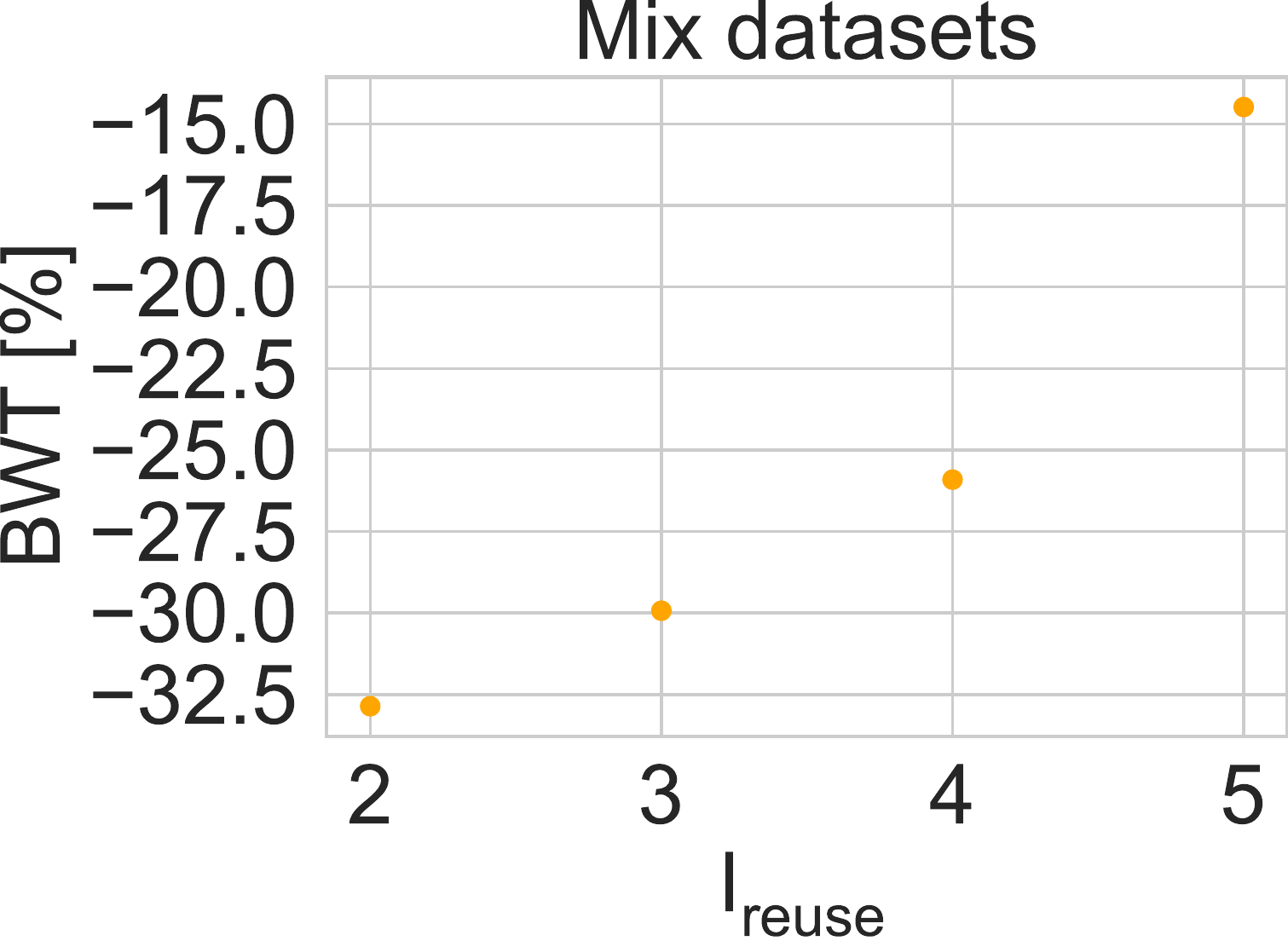} 
    \caption{BWT}
  \end{subfigure}
    \begin{subfigure}[b]{0.23\columnwidth}
    \includegraphics[width=1.\linewidth]{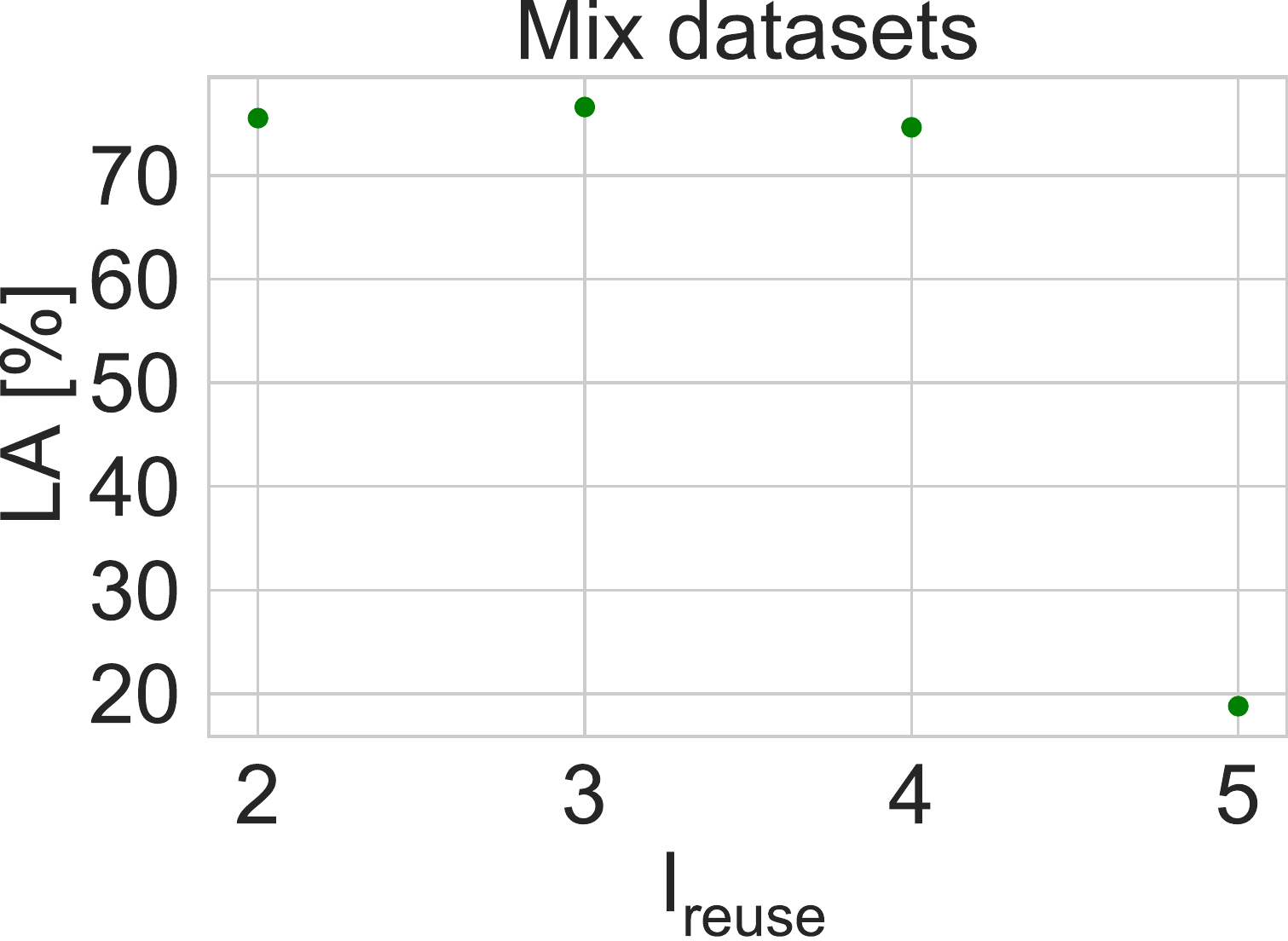} 
    \caption{LA}
  \end{subfigure}
    \begin{subfigure}[b]{0.23\columnwidth}
    \includegraphics[width=1.\linewidth]{pic/mix-lreuse_performance_.pdf} 
    \caption{Performance}
  \end{subfigure}
  \caption{Performance of AFAF at different values of $l_{reuse}$. }
  \label{fig:analysis_lreuse+detailed}
\end{figure}

In section \ref{sec:analysis}, we studied the effect of reusing full layers in learning new tasks. Figure \ref{fig:analysis_lreuse+detailed} provides the full details of the performance, including ACC, BWT, and LA. 

\begin{figure}
  \centering
  \begin{subfigure}[b]{0.23\columnwidth}
    \includegraphics[width=1\linewidth]{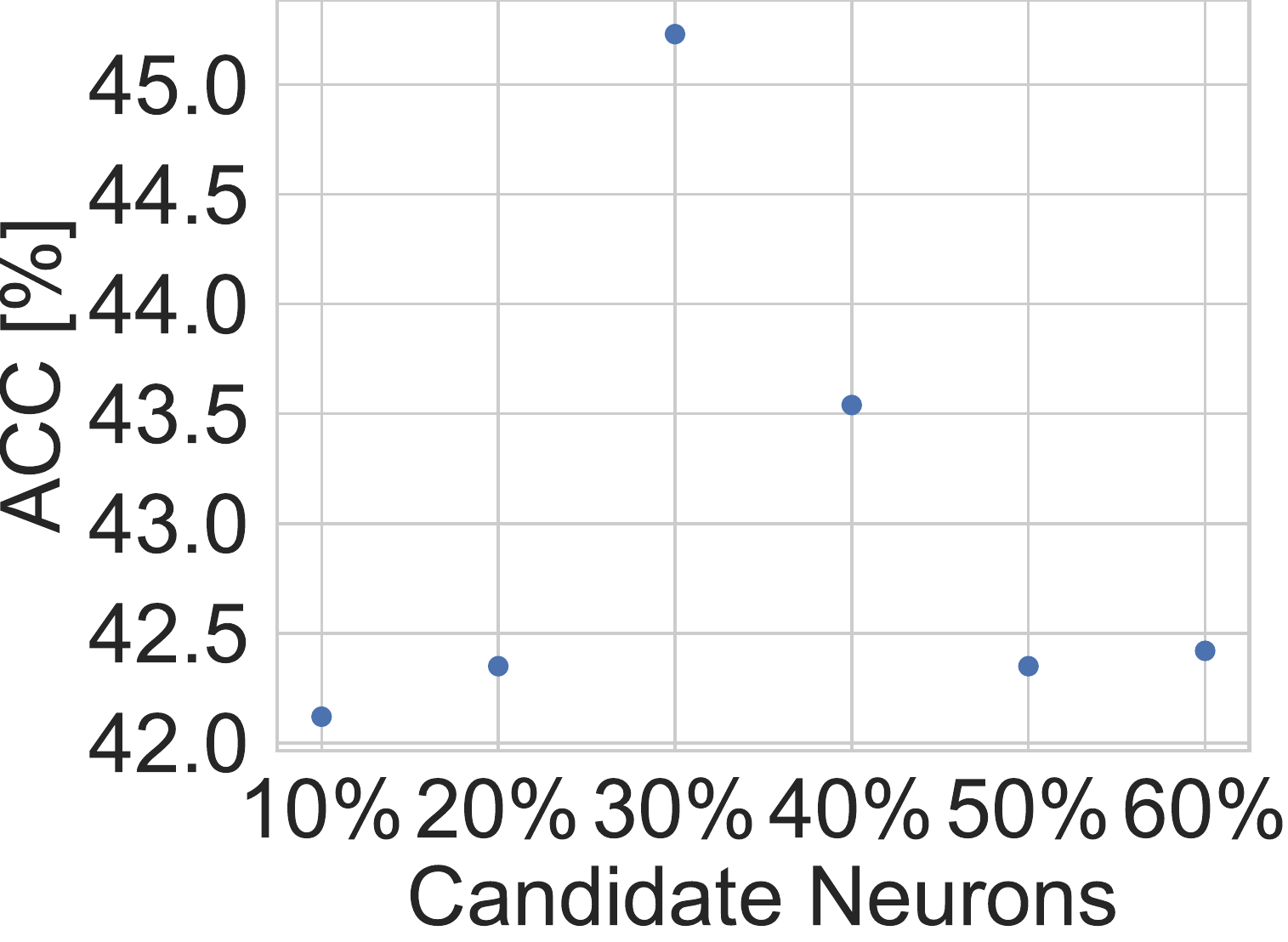} 
    \caption{ACC}
  \end{subfigure}
  \begin{subfigure}[b]{0.23\columnwidth}
    \includegraphics[width=1\linewidth]{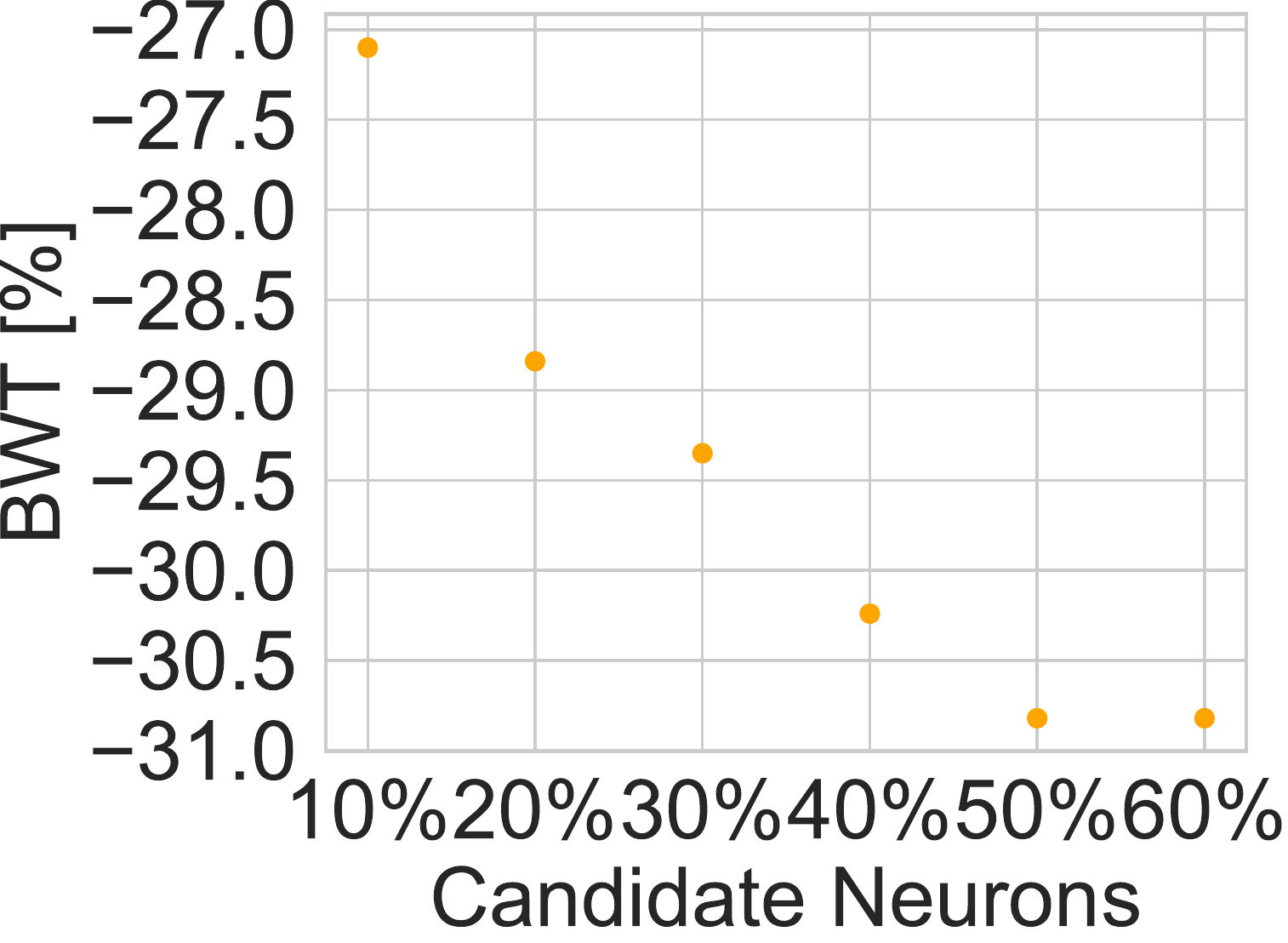} 
    \caption{BWT}
  \end{subfigure}
    \begin{subfigure}[b]{0.23\columnwidth}
    \includegraphics[width=1\linewidth]{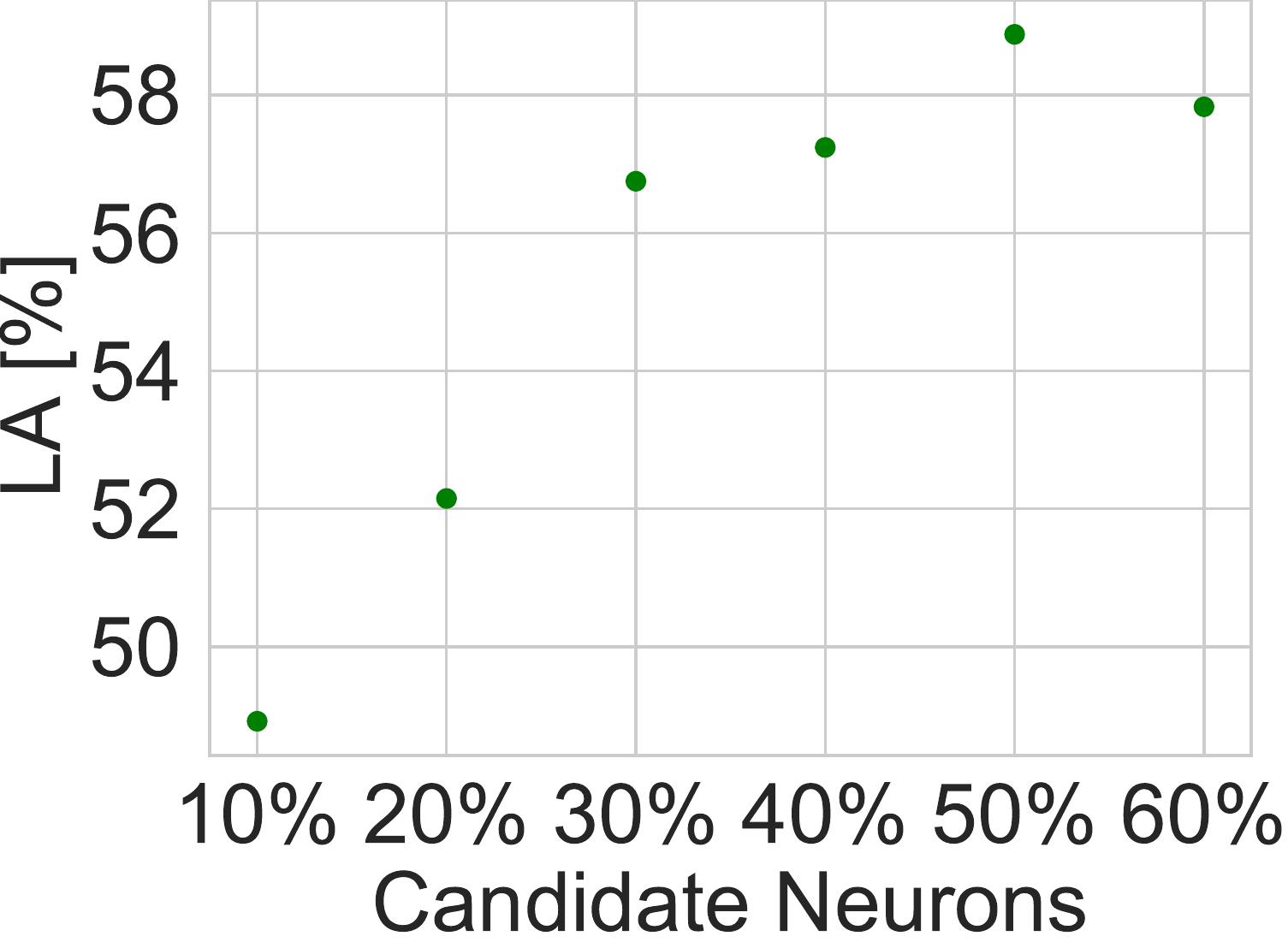} 
    \caption{LA}
  \end{subfigure}
    \begin{subfigure}[b]{0.23\columnwidth}
    \includegraphics[width=1\linewidth]{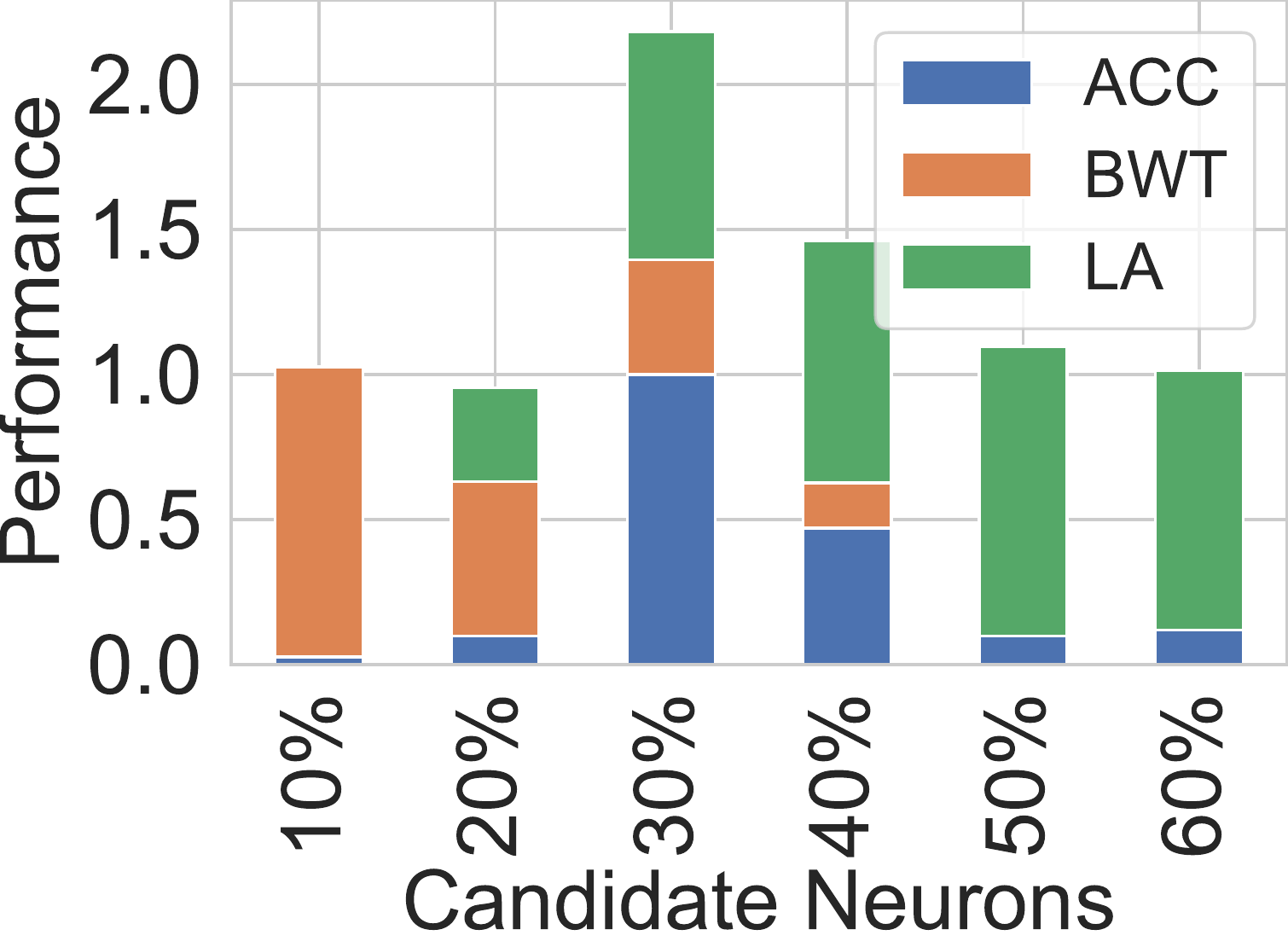} 
    \caption{Performance}
  \end{subfigure}
  \caption{Performance of AFAF at different percentages of candidate neurons $|\mathcal{R}^c_l|$ on sim-CIFAR10 benchmark. }
  \label{fig:analysis_candidate}
\end{figure}

Next, we study the effect of the number of candidate (reused) neurons on performance. The analysis is performed with $l_{reuse}=4$ (the setting from the paper). We study the performance using different percentages of the candidate neurons $\mathcal{R}_l^c$ from the allocated ones $h^{alloc}_l$ from layer $l_{reuse}$ onwards. Figure \ref{fig:analysis_candidate} shows the results on sim-CIFAR10. We observe that using a low number of candidate neurons hinders learning new tasks (i.e., low LA). Selecting $30\%\sim40\%$ of the allocated neurons based on the neuron relevance achieves the best balance of CL requirements. Using a high number of relevant neurons increases the forgetting of previous tasks since the model would be biased towards the new ones, as indicated by the higher values of LA. Please note that a higher percentage of the number of candidate neurons is not evaluated because there must be free neurons used to allocate incoming connections for the new task (Section \ref{sec:method}).

Our analyses reveal that using part of the relevant previously learned components (reusable layers/neurons) while adding new components to learn the specific representation of new tasks leads to a balance between different continual learning requirements, including: forward transfer, backward transfer, and memory and computational costs.  

\section{Representation of Similar and Dissimilar Tasks}
\label{appendix:hiddenrepresentation}
In this appendix, we visualize the representation of two sequences: one contains similar tasks while the other sequence contains dissimilar tasks. The similar sequence has two tasks from CIFAR10, where Task 1 has the two classes of \{car, cat\} while Task 2 has the two classes of \{dog, truck\}. The dissimilar sequence has the same classes in different order, where Task 1 is \{car, truck\} and Task 2 is \{cat, dog\}. We used the smaller architecture from \cite{sokar2021spacenet} for visualization purposes. 

\begin{figure}
\centering
\subfloat[\label{fig:activation_similar} Similar tasks]{\includegraphics[width=0.5\columnwidth]{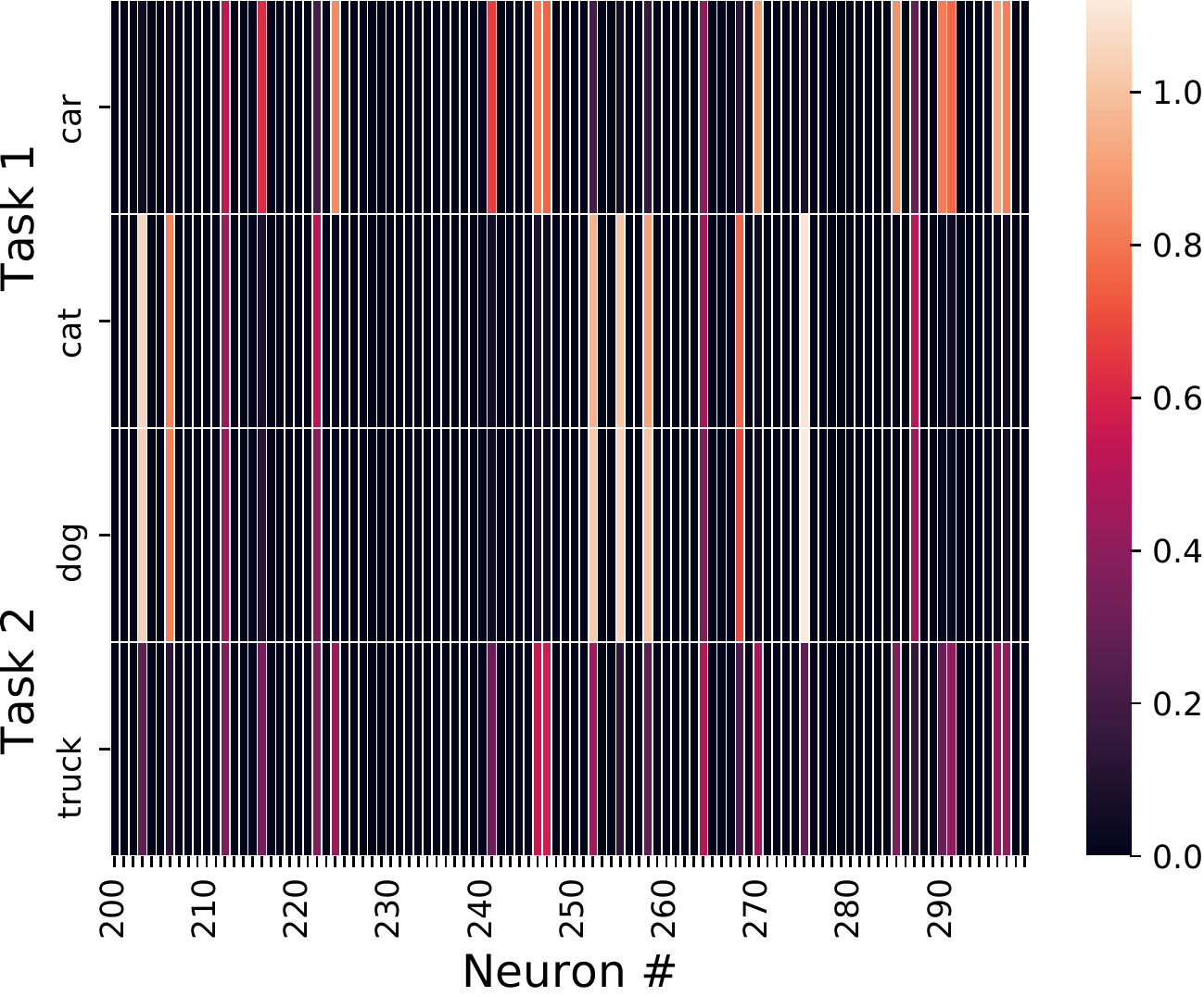}} \hfill
\subfloat[\label{fig:activation_dissimilar} Dissimilar tasks]{\includegraphics[width=0.5\columnwidth]{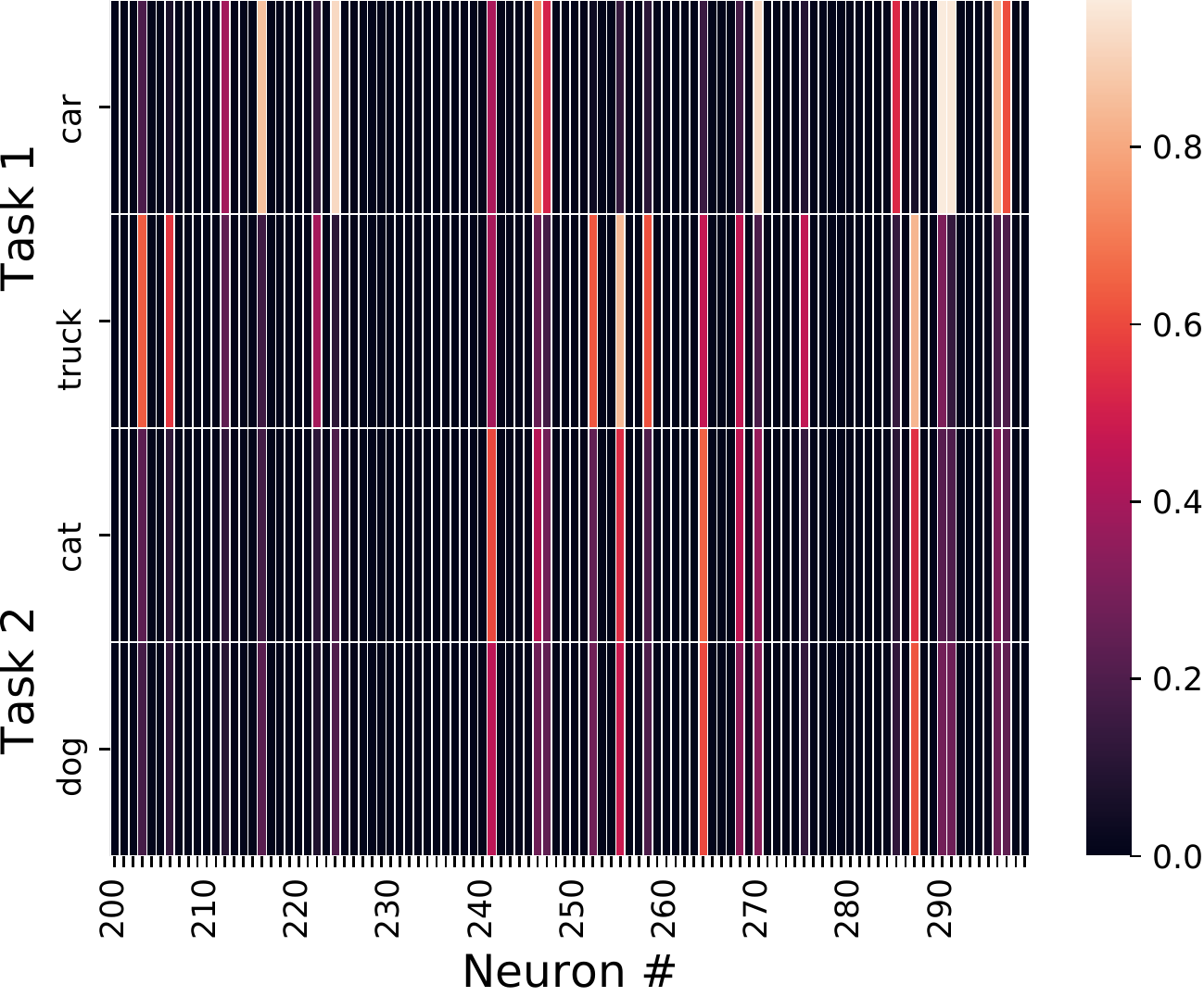}}
\caption{Visualization of the representation of a subset of the neurons in the last hidden layer of each class in similar sequence (a) and  dissimilar sequence (b).} 
\label{fig:hiddenrepresentation}
\end{figure}

Figure \ref{fig:hiddenrepresentation} shows the representations of a subset of the neurons in the last hidden layer for each sequence. We average the activation of the model trained on Task 1 over all samples of each class in Task 1 and Task 2. As illustrated in Figure \ref{fig:activation_similar}, there are high representational similarities between each class in Task 1 and the corresponding class that shares some semantic similarities in Task 2. On the other hand, if the tasks are dissimilar (Figure \ref{fig:activation_dissimilar}), the new task has a different representation from the previous one.

\section{Training Sparse Topologies}
\label{appendix:SpaceNet_DST_neuronImp}
We used the training procedure proposed in \cite{sokar2021spacenet} to train a sparse topology allocated by AFAF. The training approach is based on dynamic sparse training (DST) (Appendix \ref{appendix:DST}). The goal of this approach is to learn sparse representation for each task. During training, the sparse topology is optimized by redistributing the connections to group them in the most important neurons for the current task. The redistribution is performed through \enquote{drop-and-grow} cycles. A fraction of the least important connections is dropped, and the same fraction is added between the most important neurons. The importance of a connection is estimated based on the gradient of the loss function. Please see the full details in Algorithm 3 from \cite{sokar2021spacenet}.

\textbf{Identify neuron importance.} Following SpaceNet \cite{sokar2021spacenet}, the importance of a neuron is calculated by the summation of the importance of its connected weights. Hence, we estimate the importance of a neuron by the summation of the importance of its outgoing connections. A subset of the important neurons is fixed after training.

\section{Dynamic Sparse Training}
\label{appendix:DST}
Dynamic sparse training (DST) is a line of research that aims to reduce the computation and memory overhead of training dense neural networks by leveraging the redundancy in the parameters (i.e., being over-parametrized) \cite{denil2013predicting}. The basic idea is to train a sparse neural network from scratch and optimize the weights and topology simultaneously. Efforts in this line of research are devoted to single standard task supervised and unsupervised learning. The first work in this direction was proposed by \cite{mocanu2018scalable}. They showed that dynamic training of sparse networks from scratch achieves higher performance than dense models and static sparse neural networks trained from scratch. Recently, many interesting works have addressed this training strategy by proposing different algorithms for optimizing the topology during training \cite{hoefler2021sparsity,evci2020rigging,jayakumar2020top,NEURIPS2020_b4418237}. 

In SpaceNet \cite{sokar2021spacenet}, a DST approach was proposed to optimize the topology to produce sparse representation for continual learning. DST demonstrated its success in other fields as well, such as feature selection \cite{atashgahi2021quick}, ensembling \cite{liu2021deep}, federated learning \cite{zhu2019multi}, adversarial training \cite{ozdenizci2021training}, and deep reinforcement learning \cite{sokar2021dynamic}. 
\end{document}